\documentclass[final, 12pt]{elsarticle}
\usepackage{physics}
\usepackage[shortlabels]{enumitem}
\usepackage{mathtools}
\usepackage{amsthm}
\usepackage[utf8]{inputenc}
\usepackage{graphicx}
\graphicspath{ {images/} }
\usepackage[english]{babel}
\usepackage[margin=1in]{geometry}
\usepackage{url}
\usepackage{amsmath, nccmath}
\usepackage{float}
\usepackage{caption}
\usepackage{comment}
\usepackage{lscape}
\usepackage{afterpage}
\newcommand{\diff}{\,\text{d}}

\newcommand\restr[2]{{
		\left.\kern-\nulldelimiterspace 
		#1 
		\vphantom{\big|} 
		\right|_{#2} 
}}

\newcommand{\imp}{\mathcal{I}}
\newcommand{\pwset}{\mathcal{P}}
\newcommand{\fpwset}{\widetilde{\mathcal{P}}}

\newcommand{\tQ}{\widetilde{Q}}
\newcommand{\id}{\textbf{id}}
\newtheorem*{problem}{Open problem}

\newcommand{\fmed}{\text{m}_{\frac{1}{2}}}
\usepackage[utf8]{inputenc}
\usepackage{lmodern}
\usepackage{mathtools}
\newtheorem{thm}{Theorem}[section] 
\newtheorem{prop}[thm]{Proposition}

\newtheorem{cor}[thm]{Corollary}

\newtheorem{defn}[thm]{Definition} 
\newtheorem{example}[thm]{Example}
\usepackage{amssymb}

\usepackage{hyperref}
\journal{Fuzzy Sets and Systems}

\begin{document}

\begin{frontmatter}

    \title{Fuzzy Rough Sets Based on Fuzzy Quantification}
    \author[mymainaddress]{Adnan Theerens\corref{mycorrespondingauthor}}
    \cortext[mycorrespondingauthor]{Corresponding author}
    \ead{adnan.theerens@ugent.be}
    \author[mymainaddress]{Chris Cornelis}
    \ead{chris.cornelis@ugent.be}
    \address[mymainaddress]{Computational Web Intelligence, Department of Applied Mathematics, Computer Science and Statistics, Ghent University, Ghent, Belgium}

    \begin{abstract}
        One of the weaknesses of classical (fuzzy) rough sets is their sensitivity to noise, which is particularly undesirable for machine learning applications.
        One approach to solve this issue is by making use of fuzzy quantifiers, as done by the vaguely quantified fuzzy rough set (VQFRS) model. While this idea is intuitive, the VQFRS model suffers from both theoretical flaws as well as from suboptimal performance in applications. In this paper, we improve on VQFRS by introducing \emph{fuzzy quantifier-based fuzzy rough sets} (FQFRS), an intuitive generalization of fuzzy rough sets that makes use of general unary and binary quantification models. We show how several existing models fit in this generalization as well as how it inspires novel ones. Several binary quantification models are proposed to be used with FQFRS. We conduct a theoretical study of their properties, and investigate their potential by applying them to classification problems. In particular, we highlight Yager's Weighted Implication-based (YWI) binary quantification model, which induces a fuzzy rough set model that is both a significant
        improvement on VQFRS, as well as a worthy competitor to the popular ordered weighted averaging based fuzzy rough set (OWAFRS) model.

    \end{abstract}


\end{frontmatter}

\section{Introduction}

Fuzzy quantification is an important research topic in fuzzy logic \cite{zadehFuzzyQuantifier, yager1996quantifier, glockner2008fuzzy,delgado2014fuzzy,cascallar2020experimental}. It studies quantified sentences such as “Most Dutch people are tall” and “Nearly half of the S\&P 500 stocks are down 10\%”. Quantifiers are an effective tool to describe the quantity of elements that satisfy a certain condition. This is especially true if the condition is of a vague nature, as for example in the quantified sentence “Most Dutch people are tall”, since the quantity of elements satisfying a fuzzy condition (being tall) is hard to assess. The two most studied types of quantifiers are unary and binary quantifiers, unary quantifiers being of the form “\(Q_1\) elements are \(A\)” (e.g., “Some people are tall”) and binary quantifiers taking the form “\(Q_2\) \(A\)’s are \(B\)’s” (e.g., “Most Dutch people are tall”, assuming the universe consists of all people).
The first evaluation method for fuzzy quantified statements was introduced by Zadeh \cite{zadehFuzzyQuantifier}. His idea was to define a cardinality measure for fuzzy sets to evaluate the quantity of elements satisfying a condition. The problem with this approach is that the cardinality measure is cumulative, implying that a situation involving two people with a degree of tallness of 0.5 is regarded equivalent to one with one tall person (tallness 1) and one short person (tallness 0). An improved evaluation method was proposed by Yager \cite{yager1996quantifier}, which is based on the Ordered Weighted Averaging (OWA) operator \cite{yager1988ordered}. This method is semantically more reasonable for unary quantifiers but still lacks soundness for binary quantifiers. To resolve these issues, Glöckner \cite{glockner2008fuzzy} developed a general framework for fuzzy quantification. In this framework, fuzzy quantifiers are fully determined by how they act on classical (i.e., non-fuzzy) sets and by the choice of a quantifier fuzzification mechanism (QFM). A QFM thus reduces the evaluation of any quantified statement to the evaluation of quantified statements with crisp arguments.

An important application of fuzzy quantifiers is in (fuzzy) rough set theory.
Fuzzy rough sets (FRS, \cite{dubois1990rough}) emerge as a combination of fuzzy
sets \cite{fuzzysets} and rough sets \cite{pawlak1982rough}: while the former model vague information by recognizing
that membership to certain concepts, or logical truth of
certain propositions, is a matter of degree, the latter handle
potentially inconsistent information by providing a lower
and upper approximation of a concept with respect to an indiscernibility
relation between objects. Fuzzy rough sets extend rough sets by allowing both the concept as well as the indiscerniblity relation to be fuzzy. In rough set theory, the lower and upper approximation contain all objects
that are certainly, respectively possibly part of the concept. The condition for belonging to the lower and upper
approximation in rough sets may be expressed using fuzzy quantifiers.
For example, classically an object is a member of
the lower approximation of a concept if all objects
indiscernible from it also belong to the concept. Here,
instead of the traditional universal quantifier, one
may use a fuzzy quantifier like “most”. The purpose of
such a relaxation is to introduce a measure of tolerance
towards inconsistency into the approximations, making
them more robust, which is particularly relevant for the applications based on them \cite{vluymans2015applications}. This idea was first explored in vaguely quantified fuzzy rough sets (VQFRS, \cite{cornelis2007vaguely}) and was recently revived by the introduction of Choquet-based fuzzy rough sets (CFRS, \cite{theerens2022choquet}). It has also been shown \cite{theerens2022choquet} that Ordered Weighted Averaging (OWA) based fuzzy rough sets (OWAFRS, \cite{cornelis2010ordered}) can be interpreted naturally in terms of fuzzy quantifiers.

In this paper, we introduce a generalization of fuzzy rough sets, called fuzzy quantifier-based fuzzy rough sets (FQFRS), that takes the idea behind VQFRS and CFRS one step further. It does this by using general binary and unary quantification models to
determine the lower and upper approximation of a concept, respectively. The motivation behind this is that VQFRS and CFRS make use of inadequate fuzzy quantification models. For example, the VQFRS model is based on Zadeh's quantification approach, which is known to have several shortcomings \cite{glockner2008fuzzy}. In addition, the lower and upper approximations use the same quantification model, and as we will show, this is not a good strategy.
These flaws may deteriorate the performance and/or interpretability of applications based on them.

The remainder of this paper is structured as follows. In Section \ref{sec: prelims}, we recall the
required prerequisites for fuzzy rough sets and fuzzy quantification. In Section \ref{sec: FQFRS}, fuzzy quantifier-based fuzzy rough sets (FQFRS) are introduced and their relation with existing models is investigated. Section \ref{sec: binary quantification models} discusses different (novel) binary quantification models that can be used with FQFRS to acquire improved fuzzy rough set models. In Section \ref{sec: Experiment}, we evaluate the performance of FQFRS with different binary quantification models in the scope of a classification algorithm that uses the lower approximation, and compare them with OWAFRS.
Section
\ref{sec: Conclusion} concludes this paper and describes
opportunities for future research.

Finally, we mention that a small part of the results discussed in this paper are contained in the conference contribution \cite{theerens2022fedcsis}.

\section{Preliminaries}
\label{sec: prelims}
\subsection{Fuzzy logic}

We will denote the set of all fuzzy sets on \(X\) as \(\fpwset(X)\). Throughout this paper, we assume \(X\) is finite.

\begin{defn}
    An element \(R\) of \(\fpwset(X\times X)\) is called a fuzzy relation. A fuzzy relation \(R\) is called \emph{reflexive} if \(R(x,x)=1\) for every \(x\in X\). For an element \(y\in X\) and a fuzzy relation \(R\in \fpwset(X\times X)\), we define the \(R\)-foreset of \(y\) as the fuzzy set \(Ry(x):= R(x,y)\).
\end{defn}

For a fuzzy set \(A\in \fpwset(X)\) and \(\alpha\in[0,1]\), we will denote the \(\alpha\)-cut and strict \(\alpha\)-cut of \(A\) as follows
\[A_{\geq\alpha}=\{x\in X \,|\, A(x)\geq \alpha\},\; A_{>\alpha}=\{x\in X \,|\, A(x) >\alpha\}.\] We will denote Zadeh's sigma-count \cite{zadeh1982test} of a fuzzy set \(A\in \fpwset(X)\) as
\[\abs{A}_\Sigma := \sum_{x\in X}A(x).\]

\begin{defn}
    \hfill
    \begin{itemize}

        \item A \emph{conjunctor} is a binary operator \(\mathcal{C}:[0,1]^2\to [0,1]\) which is increasing in both arguments and satisfies \(\mathcal{C}(0,0)=\mathcal{C}(0,1)=0\) and \(\mathcal{C}(1,x)=x\) for all \(x\in[0,1]\). A \emph{t-norm} is a commutative and associative conjunctor.
        \item A \emph{disjunctor} is a binary operator \(\mathcal{D}:[0,1]^2\to [0,1]\) which is non-decreasing in both arguments and satisfies \(\mathcal{D}(1,0)=\mathcal{D}(1,1) = 1\) and \(\mathcal{D}(0,x)=x\) for all \(x\in[0,1]\). A \emph{t-conorm} is a commutative and associative disjunctor.
        \item An \emph{implicator} is a binary operator $\imp: \left[0,1\right]^2\rightarrow \left[0,1\right]$ for which $\imp(0,0)=\imp(0,1)=\imp(1,1)=1$, \(\imp(1,0)=0\) and for all $x_1,x_2,y_1,y_2$ in $ \left[0,1\right]$ the following holds:
              \begin{enumerate}
                  \item $x_1\leq x_2\Rightarrow \imp(x_1,y_1)\geq \imp(x_2,y_1)$ (non-increasing in the first argument),
                  \item $y_1\leq y_2\Rightarrow \imp(x_1,y_1)\leq \imp(x_1,y_2)$ (non-decreasing in the second argument),
              \end{enumerate}
              If \(\imp(1,x)=1\) for all \(x\in[0,1]\), \(\imp\) is called a border implicator. The Kleene-Dienes implicator is defined as \(\mathcal{I}_{KD}(x,y):=\max(1-x,y)\).
        \item A \emph{negator} is a unary operator \(\mathcal{N}:[0,1]\to [0,1]\) which is non-increasing and satisfies \(\mathcal{N}(0)=1\) and \(\mathcal{N}(1)=0\).  The standard negator \(\lnot\) is defined by \(\lnot(x):= 1-x\) for \(x\in[0,1]\).
        \item Suppose \(\mathcal{I}\) is an implicator. The function \(\mathcal{N}_\mathcal{I}\) defined by
              \[\mathcal{N}_\mathcal{I}(x)=\mathcal{I}(x,0),\;\;\forall x \in [0,1],\]
              is called the negator induced by \(\mathcal{I}\).
        \item Suppose \(\mathcal{S}\) is a t-conorm and \(\mathcal{N}\) is a negator. The mapping \[\imp(x,y)=\mathcal{S}(\mathcal{N}(x),y),\;\;\forall x,y\in [0,1],\]
              is called the S-implicator induced by \(\mathcal{S}\) and \(\mathcal{N}\).
    \end{itemize}
\end{defn}
For any binary operator \(\mathcal{O}: [0,1]^2 \to [0,1]\) we will denote its extension to fuzzy sets (i.e., \(\fpwset(X)^2 \to \fpwset(X)\)) with the same symbol, i.e.,
\[\mathcal{O}(A,B)(x) := \mathcal{O}\left(A(x), B(x)\right), \; \; \forall x \in X.\]

\subsection{Fuzzy rough sets}
Rough sets, first introduced by Pawlak \cite{pawlak1982rough}, model uncertainty that is associated with \emph{indiscernibility}. Here, indiscernibility is defined with respect to an equivalence relation, and two elements are called \emph{indiscernible} if they are in the same equivalence class. Indiscernibility arises naturally in information systems, where two elements are considered indiscernible if they are equivalent (or similar) with regard to a set of attributes.
\begin{defn}
    An \emph{information system} $\left( X, \mathcal{A}\right)$ consists of a finite non-empty set \(X\) and a non-empty set of attributes $\mathcal{A}$, where each attribute \(a\in \mathcal{A}\) is a function $a:\ X \rightarrow V_a$, with $V_a$ the set of values the attribute $a$ can take.
    A \emph{decision system} is an information system $\left( X, \mathcal{A}\cup \{d\}\right)$, where \(d\notin \mathcal{A}\) is called the \emph{decision attribute} and each \(a\in\mathcal{A}\) is called a \emph{conditional attribute}.
\end{defn}
\begin{defn}\cite{pawlak1982rough}
    \label{defn: rough set}
    Let $A$ be a subset of $X$ and \(R\) an equivalence relation on \(X\). The \emph{lower} and \emph{upper approximation} of $A$ with respect to \(R\) are defined as:
    \begin{align*}
        \underline{apr}_{R} A & = \left\{\left.x\in X\right| \left[x\right]_R \subseteq A \right\}=\left\{\left.x\in X\right| (\forall y\in X) \left((x,y)\in R \implies y\in A\right)\right\}         \\
        \overline{apr}_{R} A  & = \left\{\left.x\in X\right| \left[x\right]_R \cap A \neq \emptyset \right\}=\left\{\left.x\in X\right| (\exists y\in X) \left((x,y)\in R \land y\in A\right)\right\}.
    \end{align*}
    The pair $\left(\underline{apr}_{R} A,\overline{apr}_{R} A\right)$ is called a \emph{rough} set.
\end{defn}

For fuzzy sets and fuzzy relations, the lower and upper approximations can be extended as follows:
\begin{defn}{\cite{radzikowska2002comparative}}
    \label{defn: ICFRS}
    Given $R\in \mathcal{F}(X\times X)$ and $A\in\mathcal{F}(X)$, the \emph{lower} and \emph{upper approximation} of $A$ w.r.t.\ $R$ are defined as:
    \begin{align}
        (\underline{\text{apr}}_{R} A)(x) & = \min\limits_{y\in X} \imp(R(x,y),A(y)), \label{lower approx}       \\
        (\overline{\text{apr}}_{R} A)(x)  & = \max\limits_{y\in X} \mathcal{C}(R(x,y),A(y)),\label{upper approx}
    \end{align}
    where $\imp$ is an implicator and $\mathcal{C}$ a conjunctor.
\end{defn}
\subsection{OWA-based fuzzy rough sets}
A downside to the classical definition of lower and upper approximation in fuzzy rough set theory is their lack of robustness. The value of the membership of an element in the lower and upper approximation is fully determined by a single element because of the minimum and maximum operators in the definition. To solve this undesirable behaviour, many alternative definitions of fuzzy rough sets were introduced.  One of these is the OWA-based fuzzy rough set model \cite{cornelis2010ordered}, which has been shown to have an excellent trade-off between performance (robustness) and theoretical properties \cite{d2015comprehensive}. The Ordered Weighted Average \cite{yagerOWA} is an aggregation operator that is defined as follows:
\begin{defn}[OWA]
    \label{OWA}
    Let \(X=\{x_1,x_2,\dots,x_n\}\), \(f:X\to \mathbb{R}\) and \(\mathbf{w}=(w_1,w_2,\dots,w_n)\) be a weighting vector, i.e., \(\mathbf{w}\in [0,1]^n\) and \(\sum_{i=1}^{n}w_i=1\), then the \emph{ordered weighted average} of \(f\) with respect to \(\mathbf{w}\) is defined as
    \[\text{OWA}_{\mathbf{w}}(f):=\sum_{i=1}^n f(x_{\sigma(i)})w_i,\]
    where $\sigma$ is a permutation of \(\{1,2,\dots,n\}\) such that
    \begin{equation*}
        f(x_{\sigma(1)})\geq f(x_{\sigma(2)}) \geq\cdots\geq f(x_{\sigma(n)}).
    \end{equation*}
\end{defn}

In OWA-based fuzzy rough sets, OWA operators replace the minimum and maximum in the lower and upper approximations in classical fuzzy rough sets.
\begin{defn} \cite{cornelis2010ordered}
    \label{owafuzzyrough}
    Given $R\in\fpwset(X\times X)$, weight vectors \(\mathbf{w}_l\) and \(\mathbf{w}_u\), and \(A\in\fpwset(X)\), the \emph{OWA lower} and \emph{upper approximation} of $A$ w.r.t.\ $R$, $\mathbf{w}_l$ and \(\mathbf{w}_u\) are given by\footnote{In \cite{cornelis2010ordered}, some additional requirements were enforced on the OWA weights, but as we showed in \cite{theerens2022choquet}, these requirements do not lead to any useful additional theoretical properties, hence we omit them here.}:
    \begin{align}
        (\underline{\text{apr}}_{R,\mathbf{w}_l}A)(x) & = OWA_{\mathbf{w}_l}\left( \imp(R(x,y),A(y))\right),        \\
        (\overline{\text{apr}}_{R,\mathbf{w}_u}A)(x)  & = OWA_{\mathbf{w}_u}\left( \mathcal{C}(R(x,y),A(y))\right),
    \end{align}
    where $\imp$ is an implicator, $\mathcal{C}$ a conjunctor and \(\imp(R(x,y),A(y))\) and \(\mathcal{C}(R(x,y),A(y))\) are seen as functions in \(y\).
\end{defn}
\subsection{The Choquet integral}
The Choquet integral induces the class of all comonotone linear aggregation functions \cite{beliakov2007aggregation}.
Since we view the Choquet integral as an aggregation operator, we restrict ourselves to measures (and Choquet integrals) on finite sets. For the general setting, we refer the reader to e.g.\ \cite{wang2010generalized}.

\begin{defn}
    Let \(X\) be a finite set. A function \(\mu:\mathcal{P}(X)\to[0,1]\) is called a \emph{monotone measure} if:
    \begin{itemize}
        \item $\mu(\emptyset)=0$ and \(\mu(X)=1\),
        \item \((\forall A,B\in(\mathcal{P}(X))(A\subseteq B \implies \mu(A)\leq \mu(B))\).
    \end{itemize}
    A monotone measure is called \emph{symmetric} if \(\mu(A)=\mu(B)\) when \(\abs{A}=\abs{B}\).
\end{defn}

\begin{defn}\cite{wang2010generalized}
    \label{defn: ChoquetIntegral}
    Let $\mu$ be a monotone measure on \(X\) and \(f:X\to\mathbb{R}\) a real-valued function. The \emph{Choquet integral} of \(f\) with respect to the measure $\mu$ is defined as:
    \begin{equation*}
        \int f \diff \mu=\sum_{i=1}^{n}\mu(A^\ast_i)\cdot\left[f(x^\ast_i)-f(x^\ast_{i-1})\right],
    \end{equation*}
    where \((x^\ast_1,x^\ast_2,\dots,x^\ast_n)\) is a permutation of \(X=(x_1,x_2,\dots,x_n)\) such that
    \begin{equation*}
        f(x^\ast_1)\leq f(x^\ast_2) \leq\cdots\leq f(x^\ast_n),
    \end{equation*}
    \(A^\ast_i:=\{x^\ast_i,\dots,x^\ast_n\}\) and \(f(x^\ast_0):=0\).
\end{defn}
The following proposition provides an equivalent definition of the Choquet integral:

\begin{prop}\cite{wang2010generalized}
    \label{prop: equivalent defn Choquet}
    Let $\mu$ be a monotone measure on \(X\), \(f:X\to\mathbb{R}\) a real-valued function. Then the following holds (using the notation of Definition \ref{defn: ChoquetIntegral}):
    \begin{equation*}
        \int f \diff \mu=\sum_{i=1}^n f(x^\ast_i)\cdot\left[\mu(A^\ast_i)-\mu(A^\ast_{i+1})\right],
    \end{equation*}
    where \(\mu(A^\ast_{n+1}):=0\).
\end{prop}

We recall that OWA operators are equivalent to Choquet integrals w.r.t.\ symmetric measures. 
\begin{prop}{\cite{beliakov2007aggregation}}
	\label{ChoquetOWA}
	The Choquet integral with respect to a symmetric measure \(\mu\) is the OWA operator with weight vector \(\mathbf{w}=(w_i)_{i=1}^n=(\mu(A_{i})-\mu(A_{i-1}))_{i=1}^n\), where \(A_i\) denotes any subset with cardinality \(i\). Conversely, the OWA operator with weight vector \(\mathbf{v}=(v_i)_{i=1}^n\) is a Choquet integral with respect to the symmetric measure \(\mu\) defined as \[(\forall A\subseteq X)(\mu(A):= \sum_{i=1}^{\abs{A}}v_i).\]
\end{prop}
Note that this allows us to redefine OWAFRS as follows:
\begin{align}
    \label{eq: lower approx OWAFRS_choq}
    (\underline{\text{apr}}_{R,\mu_l}A)(y) & =\int \mathcal{I}\left(R(x,y),A(x)\right)\diff\mu_l(x), \\
    \label{eq: upper approx OWAFRS_choq}
    (\overline{\text{apr}}_{R,\mu_u}A)(y)  & =\int
    \mathcal{C}(R(x,y),A(x))\diff\mu_u(x)
\end{align}
where \(\mu_l\) and \(\mu_u\) are two symmetric measures.

\subsection{Fuzzy quantification}
Glöckner \cite{glockner2008fuzzy} defines vague quantifiers in two steps. The first step is the specification of the vague quantifier on crisp sets, i.e., to specify the ``underlying'' \emph{semi-fuzzy quantifier}. The second step is to extend this description to fuzzy arguments, i.e., to apply a \emph{quantifier fuzzification mechanism (QFM)}.

\begin{defn}{\cite{glockner2008fuzzy}}
    An \(n\)-ary semi-fuzzy quantifier on \(X\neq \emptyset\) is a mapping \(Q:(\pwset(X))^n\to [0,1]\). An \(n\)-ary fuzzy quantifier on \(X\neq \emptyset\) is a mapping \(\widetilde{Q}:\left(\fpwset(X)\right)^n\to [0,1]\). The restriction of \(\widetilde{Q}\) to crisp sets is called the underlying semi-fuzzy quantifier of \(\widetilde{Q}\), and is denoted by \(\mathcal{U}(\widetilde{Q})\).
\end{defn}

\begin{defn}{\cite{glockner2008fuzzy}}
    A QFM \(\mathcal{F}\) assigns to each semi-fuzzy quantifier \(Q: (\mathcal{P}(X))^n \to [0,1]\) a corresponding fuzzy quantifier \(\mathcal{F}(Q): \left(\fpwset(X)\right)^n\to [0,1]\).
\end{defn}

We first recall Zadeh's and Yager's traditional approaches, where they describe fuzzy quantifiers using fuzzy sets of the unit interval.

\begin{defn}{\cite{zadehFuzzyQuantifier}}
    \label{defn: RIMquantifier}
    A fuzzy set \(\Lambda\in\widetilde{\mathcal{P}}([0,1])\) is called a \emph{regular increasing monotone (RIM) quantifier} if
    \(\Lambda\) is a non-decreasing function such that \(\Lambda(0)=0\) and \(\Lambda(1)=1\).
\end{defn}

The interpretation of the RIM quantifier \(\Lambda\) is that if \(p\) is the percentage of elements for which a certain proposition \(P\) holds, then \(\Lambda(p)\) is the truth value of the quantified proposition \(\Lambda P\).

\begin{example}
    \label{exmp: S-function and universal}
    The following RIM quantifiers represent the quantifiers ``more than \(100*k \%\)'' and ``at least \(100*k \%\)'':
    \begin{align*}
        \Lambda_{>k}(p)=\left\{
        \begin{array}{ll}
            1 & \text{ if } p > k  \\
            0 & \text{ elsewhere }
        \end{array}
        \right.\;\;
        \;\Lambda_{\geq k}(p)=\left\{
        \begin{array}{ll}
            1 & \text{ if } p \geq k \\
            0 & \text{ elsewhere }
        \end{array}
        \right..
    \end{align*}
    These RIM quantifiers also include (a representation of) the universal and existential quantifier, \(\Lambda_\forall := \Lambda_{\geq 1}\) and \(\Lambda_\exists := \Lambda_{>0}\).
    Linguistic quantifiers such as ``most'' and ``some'' can be modelled using Zadeh's S-function  (\(0\leq \alpha < \beta \leq 1\)) \cite{cornelis2007vaguely}:
    \begin{align}
        \label{ZADEH_S_function}
        \Lambda_{(\alpha, \beta)}(p) & = \left\{
        \begin{array}{ll}
            0                                       & \;p\leq \alpha                               \\
            \frac{2(p-\alpha)^2}{(\beta-\alpha)^2}  & \; \alpha \leq p \leq \frac{\alpha+\beta}{2} \\
            1-\frac{2(p-\beta)^2}{(\beta-\alpha)^2} & \;  \frac{\alpha+\beta}{2}\leq p\leq \beta   \\
            1                                       & \; \beta\leq p
        \end{array}
        \right.,
    \end{align}
    for example, we could use \(\Lambda_{(0.3, 0.9)}\) and \(\Lambda_{(0.1, 0.4)}\) to model ``most'' and ``some'', respectively.
\end{example}

In Zadeh's model, unary sentences of the form ``\(\Lambda\) \(X\)'s are \(A\)'s'' and binary sentences of the form ``\(\Lambda\) \(A\)'s are \(B\)'s'', where \(\Lambda\) is a RIM quantifier and \(A,B\in\fpwset(X)\), are evaluated as

\begin{align}
    \widetilde{Z}_{\Lambda}(A)     & =\Lambda\left(\frac{\abs{A}_\Sigma}{\abs{X}}\right),              \\
    \widetilde{Z}^2_{\Lambda}(A,B) & =\Lambda\left(\frac{\abs{A\cap B}_\Sigma}{\abs{A}_\Sigma}\right),
\end{align}

respectively, while in Yager's OWA model, the unary sentence ``\(\Lambda\) \(X\)'s are \(A\)'s'' is evaluated as
\begin{equation}
    \label{yagerOWA}
    \widetilde{Y}_{\Lambda}(A) := \int A \diff \mu_\Lambda,
\end{equation}
where
\begin{equation}
    \label{eqn: WeightQuantifier}
    \mu_\Lambda(S) = \Lambda\left(\frac{\abs{S}}{\abs{X}}\right), \; \; \forall S \in \pwset(X).
\end{equation}
For the binary sentence ``\(\Lambda\) \(A\)'s are \(B\)'s'', there is no generally agreed-upon evaluation.

Note that both \(\widetilde{Z}_{\Lambda}\) and \(\widetilde{Y}_{\Lambda}\) extend the semi-fuzzy quantifier
\begin{equation}
    \label{eq: semi-fuzzy Unary Proportional}
    Q^X_{\Lambda}(S):= \Lambda \left(\frac{\abs{S}}{\abs{X}}\right), \;\; \forall S \in \pwset(X),
\end{equation}
which we will denote with \(Q_\Lambda\) if the universe \(X\) is clear from the context.

In \cite{glockner2008fuzzy}, Glöckner identified several shortcomings of Zadeh and Yager's models. To mend these flaws, he introduced so-called Determiner Fuzzification Schemes (DFS), particular QFM's that should satisfy a number of axioms\footnote{For a more in depth discussion we refer the reader to chapter three, four, and five of \cite{glockner2008fuzzy}.}.

Every QFM induces a propositional logic naturally. Indeed, note that there is a bijection between semi-fuzzy truth functions (i.e., mappings \(f: \{0,1\}^n \to [0,1]\)) and semi-fuzzy quantifiers on \(\{1,\dots,n\}\) (because of \(\{0,1\}^n \cong \pwset(\{1,\dots,n\})\)), and analogously between fuzzy truth functions (i.e., mappings \(\widetilde{f}: [0,1]^n\to[0,1]\)) and fuzzy quantifiers. Therefore, any QFM \(\mathcal{F}\) can be used to extend a semi-fuzzy truth function \(f\) to a fuzzy truth function that we will denote by \(\mathcal{F}(f)\) or \(\widetilde{f}\) if \(\mathcal{F}\) is clear from the context.
In particular, any logical operator can be extended to the unit interval (e.g.\ \(\lor\) to \(\widetilde{\lor}\)), and hence also the complement, union, and intersection (e.g.\ \(\cap\) to \(\widetilde{\cap}\)). The DFS axioms guarantee that the induced propositional logic is well-behaved:
\begin{prop}[\cite{glockner2008fuzzy}]
    If \(\mathcal{F}\) is a DFS, then \(\widetilde{\lnot}\) is a strong negator, \(\widetilde{\land}\) is a t-norm, \(\widetilde{\lor}\) is the \(\widetilde{\lnot}\)-dual t-conorm of \(\widetilde{\land}\), and \(\widetilde{\rightarrow}\) is the \(S\)-implicator induced by \(\widetilde{\lor}\) and \(\widetilde{\lnot}\).
\end{prop}

It has been shown that it is impossible for a QFM based on regular \(\alpha\)-cuts to be a DFS, because of the lack of symmetry of regular \(\alpha\)-cuts with respect to the complement. Therefore, Glöckner introduced QFM's based on three-valued cuts, since these cuts are symmetrical. We now recall some of the most important DFS's based on three-valued cuts.

\begin{defn}
    Let \(A\in\fpwset(X)\) be a fuzzy set and \(\gamma\in[0,1]\). The sets \(A^{\min}_\gamma,A^{\max}_\gamma \in \pwset(X)\) are defined by
    \begin{align*}
        A^{\min}_\gamma & =\left\{\begin{array}{cl}
            A_{\geq \frac{1}{2}\left(1+\gamma\right)} & :\; \gamma >0  \\
            A_{>\frac{1}{2}}                          & :\; \gamma = 0
        \end{array}
        \right.,\;\;A^{\max}_\gamma =\left\{\begin{array}{cl}
            A_{> \frac{1}{2}\left(1-\gamma\right)} & :\; \gamma >0  \\
            A_{\geq\frac{1}{2}}                    & :\; \gamma = 0
        \end{array}
        \right..
    \end{align*}
    The three-valued cut of \(A\) at \(\gamma\) is the three-valued subset \(A_\gamma(x):=\max(A^{\min}_\gamma(x), A^{\max}_\gamma(x))\).
\end{defn}

\begin{defn}
    The \emph{generalized fuzzy median} \(\fmed:[0,1]^2\to [0,1]\) is defined by
    \[\fmed(a,b)=\left\{
        \begin{array}{cl}
            \min(a,b) & \text{ if } \min(a,b) > 0.5 \\
            \max(a,b) & \text{ if } \max(a,b) < 0.5 \\
            0.5       & \text{ elsewhere}
        \end{array}
        \right.,\]
    for all \(a,b\in [0,1]\).
\end{defn}

\begin{defn}
    Let \(Q:(\pwset(X))^n\to [0,1]\) be a semi-fuzzy quantifier and \(A_1,\dots,A_n\in\fpwset(X)\). The functions \(\top_{Q,A_1,\dots,A_n}, \bot_{Q,A_1,\dots,A_n}:[0,1]\to [0,1]\) are defined as:
    \begin{align*}
        \top_{Q,A_1,\dots,A_n}(\gamma) & = \sup\left\{Q(B_1,\dots,B_n) \vert \left(A_i\right)_{\gamma}^{\min}\subseteq B_i \subseteq \left(A_i\right)_{\gamma}^{\max}\right\}  \\
        \bot_{Q,A_1,\dots,A_n}(\gamma) & = \inf\left\{Q(B_1,\dots,B_n) \vert \left(A_i\right)_{\gamma}^{\min}\subseteq B_i \subseteq \left(A_i\right)_{\gamma}^{\max}\right\},
    \end{align*}
    for every \(\gamma\in[0,1]\).
\end{defn}

\begin{defn}
    For every $\gamma\in [0,1]$, we define the QFM \((\cdot)_\gamma\) as
    \[Q_\gamma(A_1,A_2,\dots,A_n):=\fmed\left(\{Q(B_1,B_2,\dots,B_n)\,|\, B_i \in T_\gamma(A_i)\}\right),\]
    for all semi-fuzzy quantifiers \(Q:\pwset(X)^n \to [0,1]\).
\end{defn}

\begin{defn}
    We define the QFM's \(\mathcal{M}_{CX}\) and \(\mathcal{F}_{owa}\) as
    \begin{align*}
        \mathcal{M}_{CX}(Q)(A_1,\dots,A_n)  & := \sup \left\{\min\left(\gamma,Q_\gamma(A_1,\dots, A_n)\right)\,|\, \gamma \in [0,1]\right\},                \\
        \mathcal{F}_{owa}(Q)(A_1,\dots,A_n) & := \int_0^1\left(\frac{\top_{Q,A_1,\dots,A_n}(\gamma) + \bot_{Q,A_1,\dots,A_n}(\gamma)}{2}\right)\diff \gamma
    \end{align*}
    for all semi-fuzzy quantifiers \(Q:\pwset(X)^n \to [0,1]\) and \(A_1,\dots,A_n\in \fpwset(X)\).
\end{defn}

\begin{prop}{\cite{glockner2008fuzzy}}
    The QFM's \(\mathcal{M}_{CX}\) and \(\mathcal{F}_{owa}\) are standard DFS's, i.e., their induced negator is the standard negator and their induced disjunctor is the maximum operator.
\end{prop}

\begin{prop}{\cite{glockner2008fuzzy}}
    \label{prop: FOWA OWA}
    For every \(A\in \fpwset(X)\) and non-decreasing unary semi-fuzzy quantifier \(Q:\pwset(X)\to[0,1]\) the following holds:
    \[\mathcal{F}_{owa}(Q)(A) = \int A \diff Q,\]
    if \(Q\) is a monotone measure.
\end{prop}

\section{Fuzzy quantifier-based fuzzy rough sets}
\label{sec: FQFRS}
We start this section by providing a new view of OWAFRS from the perspective of fuzzy quantification.
To do this, we first need to relate symmetric measures to RIM quantifiers.
\begin{prop}
    \label{prop: symmetryRIM}
    Let \(\mu\) be a symmetric measure on \(X\), then there exists a RIM quantifier \(\Lambda\) such that
    \begin{equation}
        \label{RIMSymmetry}
        \mu(S) = \Lambda\left(\frac{\abs{S}}{\abs{X}}\right), \;\; \forall S\in \pwset(X).
    \end{equation}
    Conversely, if \(\Lambda\) is a RIM quantifier, Eq.\ \eqref{RIMSymmetry} yields a symmetric measure.
\end{prop}
\begin{proof}
    Let \(\mu\) be a symmetric measure and define \(\Lambda\) as follows:
    \[\Lambda(x)=\sum_{1 \leq i\leq x\cdot \abs{X}} \left(\mu_i - \mu_{i-1}\right),\]
    where \(\mu_i\) denotes the measure of a set with \(i\) elements. Because \(\mu\) is a monotone measure, \(\Lambda\) is a RIM quantifier. Finally, we check Eq.\ \eqref{RIMSymmetry}:
    \begin{align*}
        \Lambda\left(\frac{\abs{S}}{\abs{X}}\right) & =\sum_{1 \leq i\leq \abs{S}} \left(\mu_i - \mu_{i-1}\right) \\
                                                    & = \mu_{\abs{S}} - \mu_0 = \mu(A).
    \end{align*}
    The converse is trivial.
\end{proof}
Using this proposition and the definition of Yager's unary quantification model (Eq.\ \eqref{yagerOWA}), we can rewrite OWAFRS (Eq.\ \eqref{eq: lower approx OWAFRS_choq} and \eqref{eq: upper approx OWAFRS_choq}) as follows:
\begin{align}
    \label{eq: lower approx OWAFRS}
    (\underline{\text{apr}}_{R,\mu_l}A)(y) & = \widetilde{Y}_{\Lambda}\left(\imp(Ry,A)\right) \\
    \label{eq: upper approx OWAFRS}
    (\overline{\text{apr}}_{R,\mu_u}A)(y)
                                           & =\widetilde{Y}_{\Upsilon}(\mathcal{C}(Ry,A)),
\end{align}
where \(\mu_l\) and \(\mu_u\) are symmetric measures, and \(\Lambda\) and \(\Upsilon\) are their corresponding RIM quantifiers. Note that the binary quantifier defined by
\begin{equation}
    \label{eq: yager easy binary quantifier}
    \widetilde{Y}^{\imp}_{\Lambda}(A,B):=\widetilde{Y}_{\Lambda}(\imp(A,B)), \; \forall A,B \in \fpwset(X),
\end{equation}
can be viewed as an evaluation of the proposition ``\(\Lambda\) \(A\)'s are \(B\)'s''.
Generalizing this idea, we now introduce fuzzy quantifier-based fuzzy rough sets (FQFRS) by allowing general (binary for lower approximation and unary for upper approximation) quantification models.
\begin{defn}[\((\widetilde{Q}_l, \widetilde{Q}_u)\)-fuzzy rough set]
    Given a reflexive fuzzy relation \(R\in\mathcal{F}(X\times X)\), fuzzy quantifiers \(\widetilde{Q}_l:(\fpwset(X))^2\to [0,1]\) and \(\widetilde{Q}_u:\fpwset(X)\to [0,1]\), and $A\in\mathcal{F}(X)$, the \emph{lower} and \emph{upper approximation} of $A$ w.r.t.\ $R$ are given by:
    \begin{align}
        (\underline{apr}_{R,\widetilde{Q}_l}A)(y) & =\widetilde{Q}_l\left(Ry,A \right),             \\
        (\overline{apr}_{R,\widetilde{Q}_u}A)(y)  & =\widetilde{Q}_u\left(\mathcal{C}(Ry,A)\right),
    \end{align}
    where \(\mathcal{C}\) is a conjunctor.
\end{defn}
Suppose \(\widetilde{Q}_l\) and \(\widetilde{Q}_u\) represent the (linguistic) quantifiers ``almost all'' and ``some'', respectively. Then the degree of membership of an element \(y\) to the lower approximation of \(A\) is equal to the truth value of the statement ``Almost all elements similar to \(y\) are in \(A\)''. The degree of membership of \(y\) to the upper approximation is equal to the truth value of the statement ``Some elements are similar to \(y\) and are in \(A\)''. Notice the difference of the arity in the used quantification models for the lower and upper approximation. We restrict the upper approximation to unary quantification because the evaluation of the proposition ``\(Q\) elements are \(A\) and \(B\)'' is in essence a unary proposition with the fuzzy set \(\mathcal{C}(A,B)\) as the argument. This is not the case for ``\(Q\) \(A\)'s are \(B\)'', which is the underlying proposition of the lower approximation.

Let us look at some examples of fuzzy quantifier-based fuzzy rough sets.
\begin{itemize}
    \item \((\widetilde{Z}^2_{\Lambda},\widetilde{Z}_{\Upsilon})\)-FRS \\
          We start off with the model derived from the simplest quantification model, the one from Zadeh. Let \(\Lambda\) and \(\Upsilon\) be two RIM quantifiers, then the lower and upper approximation for \((\widetilde{Z}^2_{\Lambda},\widetilde{Z}_{\Upsilon})\)-fuzzy rough sets are defined as:
          \begin{align*}
              (\underline{apr}_{R, \Lambda} A)(y) & :=\widetilde{Z}^{2}_\Lambda(Ry,A) =\Lambda\left(\frac{\abs{Ry\cap A}_\Sigma}{\abs{Ry}_\Sigma}\right), \\
              (\overline{apr}_{R, \Upsilon}A)(y)  & :=\widetilde{Z}_\Upsilon (Ry\cap A)=\Upsilon\left(\frac{\abs{Ry\cap A}_\Sigma}{\abs{X}}\right).
          \end{align*}
          This closely resembles the Vaguely Quantified Fuzzy Rough Sets (VQFRS) model \cite{cornelis2007vaguely}, which uses the following lower and upper approximations:
          \begin{align*}
              (\underline{apr}^{\text{VQFRS}}_{R, \Lambda} A)(y) & :=\Lambda\left(\frac{\abs{Ry\cap A}_\Sigma}{\abs{Ry}_\Sigma}\right)=\widetilde{Z}^{2}_\Lambda(Ry,A),   \\
              (\overline{apr}^{\text{VQFRS}}_{R, \Upsilon}A)(y)  & :=\Upsilon\left(\frac{\abs{Ry\cap A}_\Sigma}{\abs{Ry}_\Sigma}\right)=\widetilde{Z}^{2}_\Upsilon(Ry,A).
          \end{align*}
          For both models, the lower approximations are identical, but whereas for VQFRS the lower and upper approximation only differ in their used RIM quantifier, \((\widetilde{Z}^2_{\Lambda},\widetilde{Z}_{\Upsilon})\)-FRS evaluates the upper approximation using Zadeh's unary quantifier \(\widetilde{Z}_{\Upsilon}\). Comparing the upper approximations of these two models, we can see that VQFRS will always be larger, since \(\abs{X}\geq \abs{Ry}\) and \(\Upsilon\) is a RIM quantifier. In some cases, the upper approximation of VQFRS becomes too large as the following example demonstrates.
          \begin{example}
              Let \(y\in X\) and \(A\in \fpwset(X)\). Suppose there are 100 elements in \(A\) with a similarity of \(0.01\) to \(y\notin A\) and the rest of the elements are not similar to \(y\) at all (\(\abs{Ry\cap A}=1\) and \(\abs{Ry}=2\)), then the upper approximation will always be 1 in the VQFRS approach:
              \[(\overline{apr}^{\text{VQFRS}}_{R, \Upsilon}A)(y)=\widetilde{Z}^2_\Upsilon(Ry,A)=\Upsilon\left(0.5\right) := 1,\]
              since \(\Upsilon\) should represent ``some''.
              Thus if one wants to discard the outlying elements from the upper approximation, this is problematic.
              In \((\widetilde{Z}^2_{\Lambda},\widetilde{Z}_{\Upsilon})\)-FRS, we get a less extreme result:
              \[(\overline{apr}_{R, \Upsilon}A)(y)=\widetilde{Z}_\Upsilon(Ry\cap A)=\Upsilon\left(\frac{1}{\abs{X}}\right).\]
          \end{example}
          Lastly we note that using the existential quantifier (i.e., \(\Upsilon = \Lambda_{\exists}\)) the upper approximations in VQFRS and in \((Z^2_\Lambda, Z_\Gamma)\)-FRS are equivalent (\(\abs{Ry}\geq 1\) since \(R\) is reflexive).
    \item \((\widetilde{Y}^{\rightarrow}_{\Lambda},\widetilde{Y}_{\Upsilon})\)-FRS \\
          As shown in Equations \eqref{eq: lower approx OWAFRS} and \eqref{eq: upper approx OWAFRS},  \((\widetilde{Y}^{\rightarrow}_{\Lambda},\widetilde{Y}_{\Upsilon})\)-FRS corresponds to OWAFRS which is preferred over VQFRS \cite{d2015comprehensive} in terms of the theoretical properties it satisfies, and in terms of robustness. Since Yager's unary model is generally accepted as a better model compared to Zadeh's, this justifies the improvement from a fuzzy quantifier perspective.
\end{itemize}

\section{Binary quantification models}
\label{sec: binary quantification models}

As we will demonstrate in the next subsection, Yager's binary quantification model has several flaws, and these flaws are handed down to OWAFRS, as evidenced by Eq.\ \eqref{eq: lower approx OWAFRS} and \eqref{eq: upper approx OWAFRS}. This is why in this section we take a look at other binary quantification models that might fix some of these shortcomings, and will thus hopefully result in a better fuzzy rough set model.

\subsection{Problem with Yager's implication-based binary quantification model / OWAFRS}
\label{sec: problem}
The following proposition and example show a flaw that OWAFRS inherits from its underlying semi-fuzzy quantifier.

\begin{defn}
    Given a RIM quantifier \(\Lambda\), we define the following semi-fuzzy quantifiers:
    \begin{align*}
        Q^2_\Lambda(A,B)             & := \Lambda\left(\frac{\abs{A\cap B}}{\abs{A}}\right)                                                                        \\
        Q^{\rightarrow}_\Lambda(A,B) & :=\Lambda\left(\frac{\abs{A\rightarrow B}}{\abs{X}}\right):=\Lambda\left(\frac{\abs{\lnot A}+\abs{A\cap B}}{\abs{X}}\right)
    \end{align*}
\end{defn}

\begin{prop}
    For all crisp sets \(A,B \in \pwset(X)\), we have
    \[\widetilde{Y}^{\imp}_{\Lambda}(A,B)=Q^{\rightarrow}_\Lambda(A,B),\]
    i.e., the underlying semi-fuzzy quantifier of \(\widetilde{Y}^{\imp}_{\Lambda}\) is \(Q^{\rightarrow}_\Lambda\).
\end{prop}
\begin{proof}
    Follows directly from the definition.
\end{proof}
Evaluating “\(\Lambda\) $Ry$ are $A$” using \(Q^{\rightarrow}_{\Lambda}\) yields:
\[\Lambda\left(\frac{\abs{Ry\rightarrow A}}{\abs{X}}\right)=\Lambda\left(\frac{\abs{\lnot Ry}+\abs{Ry\cap A}}{\abs{X}}\right).\]
Thus, the smaller the cardinality of \(Ry\), the truer the statement is. This is not what one would expect. Indeed, let \(y\notin A\) be an instance that is an outlier (\(Ry\) only contains \(y\)), then the membership of \(y\) to the lower approximation of \(A\) is always very high, regardless of \(A\).
The problem with \(Q^{\rightarrow}_\Lambda\) is that it evaluates ``For most elements of \(X\), if they are in \(A\), they are in \(B\)'' instead of ``Most \(A\)'s are \(B\)'s''. In the first one, all elements matter, while for the second one only elements of \(A\) matter. The following example demonstrates how important this difference is.

\begin{example}
    \label{exmp: issueImplicator}
    Let us look at the difference between ``Most Belgian people are not Belgian'' and ``For most people in the world, if they are Belgian, they are not Belgian''. Most people would agree that both sentences are plainly wrong. But if we evaluate the first sentence using \(Q_\Lambda\) and the second one using \(Q^{\rightarrow}_\Lambda\), we get the following:
    \begin{align*}
        Q^{2}_\Lambda(\text{Belgian}, \lnot \text{Belgian})           & = \Lambda\left(\frac{\abs{\emptyset}}{\abs{Belgian}}\right)= 0,                        \\
        Q^{\rightarrow}_\Lambda(\text{Belgian}, \lnot \text{Belgian}) & = \Lambda\left(\frac{\abs{\lnot \text{Belgian}}}{\abs{\text{World}}}\right) \approx 1,
    \end{align*}
    because the percentage of Belgians in the world is minuscule. So the second one is still correct since most people are simply not from Belgium.
\end{example}
A more sensible underlying semi-fuzzy quantifier is \(Q^2_{\Lambda}\), since this is exactly how one would evaluate ``Most \(A\)'s are \(B\)'s'' when \(A\) and \(B\) are crisp sets. Indeed, when \(A\) and \(B\) are crisp, we can easily calculate the percentage of elements of \(A\) that are in \(B\), which after applying the RIM quantifier \(\Lambda\) (cf.\ Definition \ref{defn: RIMquantifier} and below) gives us \(Q^2_\Lambda\). One fuzzy quantifier that has \(Q^2_{\Lambda}\) as its underlying semi-fuzzy quantifier is Zadeh's \(\widetilde{Z}^2_{\Lambda}\), but this quantifier of course has some other issues. This is why in the next section we consider other fuzzy quantifiers that extend \(Q^2_{\Lambda}\).

\subsection{Solutions}
\label{sec: solutions}

\subsubsection{Solution 1: DFS-based binary quantification models}
The most obvious solution is to start off with the desired semi-fuzzy quantifier \(Q^2_{\Lambda}\) and extend it using a DFS \(\mathcal{F}\). One choice that certainly makes sense is \(\mathcal{F}_{owa}\), which generalizes Yager's OWA approach \cite{yager1996quantifier}.
\begin{prop}
    For every RIM quantifier \(\Lambda\) we have
    \[\mathcal{F}_{owa}(Q_\Lambda) = \widetilde{Y}_{\Lambda}.\]
\end{prop}
\begin{proof}
    Follows from Proposition \ref{prop: FOWA OWA} and the fact that \(Q_\Lambda = \mu_\Lambda\) (Eq.\ \eqref{eq: semi-fuzzy Unary Proportional}).
\end{proof}

Another option is \(\mathcal{M}_{CX}\), which is known to be very interesting from a theoretical perspective, but due to the coarse granularity of truth values and cautiousness might not be the best choice for fuzzy rough sets (cf.\ Section 7.13 in \cite{glockner2008fuzzy}), when applied to a classification problem. Indeed, if the model is too cautious (i.e., evaluates sentences as \(0.5\) most of the time) or there are not many options for the truth values, the membership degree to the lower approximation of many classes will be the same, leaving a classification algorithm indecisive. 

A downside to the above solutions is that they are computationally more complex than \(\widetilde{Y}^{\imp}_{\Lambda}\) (\(\mathcal{O}(\abs{X}^2)\) vs.\ \(\mathcal{O}(\abs{X}\log(\abs{X}))\)), as well as being more complex to implement (cf.\ Chapter 11 in \cite{glockner2008fuzzy}).

\subsubsection{Solution 2: Weighted Ordered Weighted Average-based binary quantification model}

Looking at Yager's model \(\widetilde{Y}^{\imp}_{\Lambda}\), we can see that an element outside \(A\) contributes as much to the truth value as an element that is both in \(A\) and in \(B\). Therefore we now introduce a new binary quantification model that applies an extra weighting to elements of \(A\) to compensate for this issue:

\begin{defn}
    Let \(\Lambda\) be a RIM quantifier, \(\imp\) an implicator and \(A,B\in\fpwset(X)\). We define the fuzzy quantifier \(\widetilde{W}^{\imp}_\Lambda: \fpwset(X)\to [0,1]\) as:
    \begin{align*}
        \widetilde{W}^{\imp}_\Lambda(A,B) & :=\int \imp(A,B)\diff \mu^A_\Lambda,\;\;\mu^A_\Lambda(S):=\Lambda\left(\frac{\abs{S\cap A}_\Sigma}{\abs{A}_\Sigma}\right),
    \end{align*}
    where \(S\in \pwset(X)\).
\end{defn}

The following proposition shows that \(\widetilde{W}^{\imp}_\Lambda\) is based on a Weighted Ordered Weighted Averaging (WOWA) operator \cite{WOWA} that uses \(\Lambda\) for the OWA part and assigns to every element \(x\in X\) the weight \(p_x = A(x)/\abs{A}\).

\begin{prop}
    \label{prop: yager vs wowa}
    Let \(\Lambda\) be a RIM quantifier, \(\imp\) an implicator, \(A,B\in \fpwset(X)\). Now define \(x^\ast_i\) such that \(\imp(A,B)(x^{\ast}_i)\) is the \(i\)th largest value of \(\imp(A,B)(x)\) for all \(x\in X\) and \(i \in \{1,\dots,\abs{X}=n\}\). Then we have the following:
    \begin{align}
        \label{eq: WOWA binary quantifier}
        \widetilde{W}^{\imp}_\Lambda (A,B) & = \sum_{i=1}^n (\imp(A,B))(x^{\ast}_i)\cdot \left(\Lambda\left(\frac{\sum_{j=1}^i A(x^{\ast}_j)}{\abs{A}_{\Sigma}}\right)-\Lambda\left(\frac{\sum_{j=1}^{i-1} A(x^{\ast}_j)}{\abs{A}_{\Sigma}}\right)\right).
    \end{align}
\end{prop}
\begin{proof}
    To prove Equation \eqref{eq: WOWA binary quantifier}, let us rewrite the Choquet integral in \(\widetilde{W}^{\imp}_{\Lambda}\) using Proposition \ref{prop: equivalent defn Choquet}:
    \begin{align*}
        \widetilde{W}^{\imp}_\Lambda (A,B) & = \sum_{i=1}^n (\imp(A,B))(x^{\ast}_{n-i+1})\cdot \left(\mu^A_{\Lambda}\left(\{x^\ast_{1},\dots,x^\ast_{n-i+1}\}\right)-\mu^A_{\Lambda}\left(\{x^\ast_{1},\dots,x^\ast_{n-i}\}\right)\right)                           \\
                                           & = \sum_{i=1}^n (\imp(A,B))(x^{\ast}_{n-i+1})\cdot \left(\Lambda\left(\frac{\sum_{j=1}^{n-i+1}A(x^{\ast}_j)}{\abs{A}_{\Sigma}}\right)-\Lambda\left(\frac{\sum_{j=1}^{n-i}A(x^{\ast}_j)}{\abs{A}_{\Sigma}}\right)\right) \\
                                           & = \sum_{i=1}^n (\imp(A,B))(x^{\ast}_i)\cdot \left(\Lambda\left(\frac{\sum_{j=1}^i A(x^{\ast}_j)}{\abs{A}_{\Sigma}}\right)-\Lambda\left(\frac{\sum_{j=1}^{i-1} A(x^{\ast}_j)}{\abs{A}_{\Sigma}}\right)\right).
    \end{align*}
\end{proof}

The following proposition shows that this weighting indeed gives us a fuzzy quantifier with the desired underlying semi-fuzzy quantifier \({Q}^2_\Lambda\).

\begin{prop}
    For all crisp sets \(A,B \in \pwset(X)\), we have
    \[\widetilde{W}^{\imp}_{\Lambda}(A,B)=Q^{2}_\Lambda(A,B)= \Lambda\left(\frac{\abs{A\cap B}}{\abs{A}}\right),\]
    i.e.,\ \(\mathcal{U}\left(\widetilde{W}^{\imp}_\Lambda\right) = {Q}^2_\Lambda\).
\end{prop}

\begin{proof}
    Let \(A,B \in \pwset(X)\) be two crisp sets, then:
    \begin{align*}
        \widetilde{W}^{\imp}_\Lambda (A,B) & = \int \imp(A,B)\diff \mu^A_\Lambda=\mu^A_\Lambda (\imp(A,B))          \\
                                           & =\mu^A_\Lambda\left(\lnot A \cup B\right)                              \\
                                           & =\Lambda\left(\frac{\abs{A\cap B}}{\abs{A}}\right) = Q^2_\Lambda(A,B).
    \end{align*}
\end{proof}

Finally, we give a numerical example to demonstrate the WOWA-based quantification model, as well as compare it to \(\widetilde{Y}^\imp_\Lambda\).

\begin{example}
    \label{exmp: WOWABinary}
    Let \(X=\{x_1,x_2,x_3,x_4,x_5,x_6\}\), \(B = \{x_1,x_2\}\) and
    \[A= \{(x_1,1),(x_2,0.2),(x_3,0), (x_4,0),(x_5,0),(x_6,0.3)\},\]
    then we have
    \[\imp_{KD}(A,B)= \{(x_1,1),(x_2,1),(x_3,1), (x_4,1),(x_5,1),(x_6,0.7)\}.\]
    Evaluating \(\widetilde{Y}^{\imp_{KD}}_\Lambda(A,B)\) with \(\Lambda=\Lambda_{(0.7,1)}\) from Eq.\ \eqref{ZADEH_S_function}, we get

    \begin{align*}
        \widetilde{Y}^{\imp_{KD}}_\Lambda(A,B) & = \sum_{i=1}^6 (\imp_{KD}(A,B))(x^{\ast}_i)\cdot \left(\Lambda\left(\frac{i}{6}\right)-\Lambda\left(\frac{i-1}{6}\right)\right)                     \\
                                               & = 0.7 \cdot \left(\Lambda(1) - \Lambda\left(\frac{5}{6}\right)\right) + 1\cdot \Lambda\left(\frac{5}{6}\right)\approx 0.7 + 0.3\cdot \Lambda(0.833) \\
                                               & \approx 0.82,
    \end{align*}
    while for \(\widetilde{W}^{\imp_{KD}}_\Lambda(A,B)\), we get
    \begin{align*}
        \widetilde{W}^{\imp_{KD}}_\Lambda(A,B) & = \sum_{i=1}^6 (\imp_{KD}(A,B))(x^{\ast}_i)\cdot \left(\Lambda\left(\frac{\sum_{j=1}^i A(x^{\ast}_j)}{\abs{A}_{\Sigma}}\right)-\Lambda\left(\frac{\sum_{j=1}^{i-1} A(x^{\ast}_j)}{\abs{A}_{\Sigma}}\right)\right) \\
                                               & = \sum_{i=1}^6 (\imp_{KD}(A,B))(x_i)\cdot \left(\Lambda\left(\frac{\sum_{j=1}^i A(x_j)}{1.5}\right)-\Lambda\left(\frac{\sum_{j=1}^{i-1} A(x_j)}{1.5}\right)\right)                                                \\
                                               & = 0.7 \cdot \left(\Lambda(1) - \Lambda\left(\frac{1.2}{1.5}\right)\right) + 1\cdot \Lambda\left(\frac{1.2}{1.5}\right)=0.7 + 0.3\cdot \Lambda(0.8)                                                                \\
                                               & \approx 0.77.
    \end{align*}
    Note that \(\widetilde{W}^\imp_\Lambda\) indeed gives less weight to all the instances for which the membership degree to \(A\) is zero.
\end{example}

\subsubsection{Solution 3: Yager's weighted implication-based binary quantification model}
Instead of the WOWA approach for the weighting, we also consider another solution. We will call it Yager's Weighted Implication (YWI) based quantification model, reflecting the fact that it is a generalization of a model proposed by Yager in \cite{yager1991fuzzy}.
\begin{defn}
    Let \(\Lambda\) be a RIM quantifier, \(\imp\) an implicator and \(A,B\in\fpwset(X)\). We define the fuzzy quantifier \(\widetilde{Y}^{2}_\Lambda: \fpwset(X)\to [0,1]\) as:
    \begin{align*}
        \widetilde{Y}^{2}_\Lambda(A,B) & :=\int \imp(A,B)\diff \mu'_\Lambda,\;\;\mu'_\Lambda(S):=\Lambda\left(\frac{\sum_{j=1}^{\abs{S}}A(x^{\ast}_{j})}{\abs{A}_\Sigma}\right),
    \end{align*}
    where \(S\in \pwset(X)\) and \(A(x^\ast_i)\) is the \(i\)th smallest value of \(A(x)\) for \(x\in X\) and \(S\in\pwset(X)\).
\end{defn}
Noting that \(\mu'_\Lambda\) is a symmetric measure (thus, it induces an OWA operator) in the previous definition we can also write \(\widetilde{Y}^2_\Lambda\) as follows:
\begin{equation}
    \label{eq: yager binary quantifier}
    \widetilde{Y}^2_\Lambda (A,B)  = \sum_{i=1}^n (\imp(A,B))(x^{\ast}_i)\cdot \left(\Lambda\left(\frac{\sum_{j=1}^i A(y^{\ast}_j)}{\abs{A}_\Sigma}\right)-\Lambda\left(\frac{\sum_{j=1}^{i-1} A(y^{\ast}_j)}{\abs{A}_\Sigma}\right)\right),
\end{equation}
where \(x^\ast_i\) and \(y^\ast_i\) are defined such that \(\imp(A,B)(x^{\ast}_i)\) is the \(i\)th largest value of \(\imp(A,B)(x)\) and \(A(y^\ast_i)\) is the \(i\)th smallest value of \(A(x)\) for all \(x\in X\) and \(i \in \{1,\dots,\abs{X}=n\}\).

The following proposition shows that this weighting indeed gives us a fuzzy quantifier with the desired underlying semi-fuzzy quantifier \({Q}^2_\Lambda\).
\begin{prop}
    For all crisp sets \(A,B \in \pwset(X)\), we have
    \[\widetilde{Y}^{2}_{\Lambda}(A,B)=Q^{2}_\Lambda(A,B)= \Lambda\left(\frac{\abs{A\cap B}}{\abs{A}}\right),\]
    i.e., \(\mathcal{U}\left(\widetilde{Y}^{2}_\Lambda\right) = {Q}^2_\Lambda\).
\end{prop}
\begin{proof}
    Let \(A,B \in \pwset(X)\), then:
    \begin{align*}
        \widetilde{Y}^2_{\Lambda}(A,B) & = \int \imp (A,B)\diff \mu'_\Lambda = \mu'_\Lambda(\imp(A,B)) \\
                                       & =\mu'_\Lambda\left(\lnot A \cup B\right)                      \\
                                       & =\mu'_\Lambda\left(\lnot A \cup (A\cap B)\right)
    \end{align*}
    But for crisp sets \(A\), the measure \(\mu'_\Lambda\) reduces to:
    \[\mu'_\Lambda(S)=\left\{
        \begin{array}{cl}
            0                                     & \text{ if } \abs{S} \leq \abs{\lnot A} \\
            \frac{\abs{S}-\abs{\lnot A}}{\abs{A}} & \text{ if } \abs{S} > \abs{\lnot A}
        \end{array}
        \right.,\]
    from which we get the desired result:
    \begin{align*}
        \widetilde{Y}^2_{\Lambda}(A,B) & =\mu'_\Lambda\left(\lnot A \cup (A\cap B)\right)=\Lambda\left(\frac{\abs{\lnot A \cup (A\cap B)}-\abs{\lnot A}}{\abs{A}}\right) \\
                                       & = \Lambda\left(\frac{\abs{A\cap B}}{\abs{A}}\right).
    \end{align*}
\end{proof}
Finally, we give a numerical example for \(\widetilde{Y}^{2}_{\Lambda}\).
\begin{example}
    Recall Example \ref{exmp: WOWABinary}, and let us evaluate \(\widetilde{Y}^{2}_{\Lambda}(A,B)\):
    \begin{align*}
        \widetilde{Y}^{2}_\Lambda(A,B) & = \sum_{i=1}^6 (\imp_{KD}(A,B))(x^{\ast}_i)\cdot \left(\Lambda\left(\frac{\sum_{j=1}^i A(y^{\ast}_j)}{\abs{A}_{\Sigma}}\right)-\Lambda\left(\frac{\sum_{j=1}^{i-1} A(y^{\ast}_j)}{\abs{A}_{\Sigma}}\right)\right) \\
                                       & = \sum_{i=1}^6 (\imp_{KD}(A,B))(x_i)\cdot \left(\Lambda\left(\frac{\sum_{j=1}^i A(y^{\ast}_j)}{1.5}\right)-\Lambda\left(\frac{\sum_{j=1}^{i-1} A(y^{\ast}_j)}{1.5}\right)\right)                                  \\
                                       & = 0.7 \cdot \left(\Lambda(1) - \Lambda\left(\frac{0.5}{1.5}\right)\right) + 1\cdot \Lambda\left(\frac{0.5}{1.5}\right)=0.7 + 0.3\cdot \Lambda(0.33)                                                               \\
                                       & = 0.7.
    \end{align*}
    Note that compared to \(\widetilde{W}^{\imp}_{\Lambda}\), \(\widetilde{Y}^{2}_{\Lambda}(A,B)\) performs a more drastic weighting on \(A\), resulting in a smaller truth value for the proposition ``Most \(A\)'s are \(B\)'s''.
\end{example}

\subsection{Theoretical study of the binary quantification models}
In this subsection, we conduct a theoretical study of the different proposed binary quantification models, in order to shed some light on which solution is more preferable.
\subsubsection{Yager's implication based binary quantification model linked to DFS}
We start off by showing that we can view \(\widetilde{Y}^{\imp_{KD}}_\Lambda\), a specific instantiation of the binary fuzzy quantifier from Eq.\ \eqref{eq: yager easy binary quantifier} used in OWAFRS, as a DFS-based model.
To do this, we show that evaluating \(\mathcal{F}(Q^{\rightarrow}_{\Lambda})(A,B)\) for fuzzy sets \(A,B\) and a DFS \(\mathcal{F}\) simply amounts to evaluating the fuzzy set \(A\tilde{\rightarrow}B\) using the unary quantifier \(\mathcal{F}(Q_{\Lambda})\), where \(\tilde{\rightarrow}\) is the implicator induced by the DFS \(\mathcal{F}\).
\begin{defn}
    Let \(\tQ:(\fpwset(X))^n\to [0,1]\) be a fuzzy quantifier, then the fuzzy quantifier \(\tQ\tilde{\rightarrow}: (\fpwset(X))^{n+1}\to [0,1]\) is defined as:
    \[\tQ\tilde{\rightarrow}(A_1,\dots,A_{n+1}):=\tQ(A_1,\dots,A_{n-1},(A_n\tilde{\rightarrow} A_{n+1})).\]
    For a semi-fuzzy quantifier \(Q\), the semi-fuzzy quantifier \(Q\rightarrow\) is defined analogously.
\end{defn}
\begin{prop}
    For every semi-fuzzy quantifier \(Q\) and DFS \(\mathcal{F}\) we have:
    \[\mathcal{F}(Q\rightarrow)=\mathcal{F}(Q)\tilde{\rightarrow}.\]
\end{prop}
\begin{proof}
    This follows from the fact that a DFS is compatible with internal meets and internal negations \cite{glockner2008fuzzy}.
\end{proof}
\begin{cor}
    \label{cor: implicative quantifier}
    Let \(\mathcal{F}\) be a DFS and \(Q_\Lambda\) the unary quantifier from Equation \eqref{eq: semi-fuzzy Unary Proportional}, then:
    \[\mathcal{F}\left(Q^{\rightarrow}_{\Lambda}\right)(A,B)=\mathcal{F}(Q_{\Lambda})(A\tilde{\rightarrow} B),\]
    for every \(A,B\in \fpwset(X)\).
\end{cor}
Applying this to \(\mathcal{F}_{owa}\), we can write \(\widetilde{Y}^{\imp_{KD}}_\Lambda\) as a DFS-based model:
\begin{cor}
    \label{cor: FOWA YAGER}
    \[\mathcal{F}_{owa}(Q^{\rightarrow}_\Lambda)(A,B)=\int \imp_{KD}(A,B)\diff \mu_\Lambda= \widetilde{Y}^{\imp_{KD}}_\Lambda(A,B),\]
    for every \(A,B\in\fpwset(X)\).
\end{cor}
\begin{proof}
    Follows from the fact that \(\mathcal{F}_{owa}\) is a standard DFS (thus the induced implicator is \(\imp_{KD}\)) \cite{glockner2008fuzzy} and Proposition \ref{prop: FOWA OWA}.
\end{proof}
\begin{problem}
Does there exist a DFS \(\mathcal{F}_\imp\) for every S-implicator \(\imp\) such that 
\[\mathcal{F}_\imp(Q^\rightarrow_\Lambda)(A,B) = \int \imp(A,B)\diff \mu_\Lambda?\]
Or equivalently, using Corollary \ref{cor: implicative quantifier}, does there exist a DFS \(\mathcal{F}_\imp\) for which the induced implicator is \(\imp\) and 
\[\mathcal{F}_\imp(Q)(A) = \int A \diff Q, \;\;\forall A \in \fpwset(X),\]
for every symmetric measure \(Q\) on \(X\)?
\end{problem}

\subsubsection{Behaviour under the existential and universal RIM quantifier}
When using the universal RIM quantifier \(\Lambda_\forall\) (Example \ref{exmp: S-function and universal}), we expect that the different binary quantification models reduce to the standard inclusion measure \cite{bandler1980fuzzy} based on the used implicator. The two following propositions show exactly that.
\begin{prop}
    \label{universalALL}
    For every \(A,B \in \fpwset(X)\), we have
    \begin{align*}
        \widetilde{Y}^{2}_{\Lambda_\forall}(A,B) & =\min_{x \in X} \left(\imp \left(A(x),B(x)\right)\right)= \widetilde{W}^{\imp}_{\Lambda_\forall}(A,B), \\
                                                 & =\widetilde{Y}^{\imp}_{\Lambda_\forall}(A,B).
    \end{align*}
\end{prop}
\begin{proof}
    Recall that \(\Lambda_{\forall}(x)\) is only equal to one if \(x = 1\) and is equal to zero in all other cases. Combining this with Equation \eqref{eq: yager binary quantifier} and the fact that \(A(y^\ast_i)\) is the \(i\)th smallest value of \(A(x)\), we get the first equality. For the second equality, suppose it does not hold, i.e.,
    \begin{align*}
        \widetilde{W}^{\imp}_{\Lambda_\forall}(A,B) & = \sum_{i=1}^n (\imp(A,B))(x^{\ast}_i)\cdot \left(\Lambda_\forall\left(\frac{\sum_{j=1}^i A(x^{\ast}_j)}{\abs{A}_{\Sigma}}\right)-\Lambda_\forall\left(\frac{\sum_{j=1}^{i-1} A(x^{\ast}_j)}{\abs{A}_{\Sigma}}\right)\right), \\
                                                    & \neq \imp(A,B)(x^\ast_n),
    \end{align*}
    where \(x^\ast_i\) is defined such that \(\imp(A,B)(x^{\ast}_i)\) is the \(i\)th largest value of \(\imp(A,B)(x)\) for all \(x\in X\) and \(i \in \{1,\dots,\abs{X}=n\}\).  
    Since
    \[v_i = \Lambda_\forall\left(\frac{\sum_{j=1}^i A(x^{\ast}_j)}{\abs{A}_{\Sigma}}\right)-\Lambda_\forall\left(\frac{\sum_{j=1}^{i-1} A(x^{\ast}_j)}{\abs{A}_{\Sigma}}\right)\]
    forms a weight vector and can only be equal to one or zero, we must have that
    \begin{align*}
         & v_n=0,                                                                                                     \\
         & \Longleftrightarrow \Lambda_\forall\left(\frac{\sum_{j=1}^{n-1} A(x^{\ast}_j)}{\abs{A}_{\Sigma}}\right)=1, \\
         & \Longrightarrow A(x^\ast_n)= 0,                                                                            \\
         & \Longrightarrow \imp(A,B)(x^\ast_n)=1,                                                                     \\
         & \Longrightarrow (\forall x\in X)(\imp(A,B)(x)=1),
    \end{align*}
    which leads to a contradiction
    \[\widetilde{W}^{\imp}_{\Lambda_\forall}(A,B)=1=\imp(A,B)(x^\ast_n)\neq\widetilde{W}^{\imp}_{\Lambda_\forall}(A,B).\]
    The third equality is trivial.
\end{proof}
\begin{prop}{\cite{glockner2008fuzzy}}
    \label{UniveralALLGlock}
    For any standard DFS \(\mathcal{F}\), and \(A,B \in \fpwset(X)\), we have
    \[\mathcal{F}\left(Q^2_{\Lambda_\forall}\right)(A,B)= \min_{x\in X}\left(\imp_{KD}(A(x),B(x))\right).\]
\end{prop}
Thus, when using the Kleene-Dienes implicator, \(\widetilde{Y}^{\imp}_{\Lambda_\forall}, \widetilde{W}^{\imp}_{\Lambda_\forall}, \widetilde{Y}^{2}_{\Lambda_\forall}\) and \(\mathcal{F}\left(Q^2_{\Lambda_\forall}\right)\) are all the same and the lower approximations of the FQFRS that are based on them reduce to the classical fuzzy rough lower approximation. For the existential RIM quantifier \(\Lambda_\exists\) (Example \ref{exmp: S-function and universal}) we have the following proposition.
\begin{prop}
    For every \(A,B \in \fpwset(X)\), we have
    \begin{align*}
        \widetilde{Y}^{\imp}_{\Lambda_\exists}(A,B) & =\max_{x\in X} \left(\imp (A,B)(x)\right),                                                  \\
        \widetilde{Y}^2_{\Lambda_\exists}(A,B)      & =\max_{x\in A_{>0}}\left(\imp (A,B)(x) \right)=\widetilde{W}^{\imp}_{\Lambda_\exists}(A,B).
    \end{align*}
\end{prop}
\begin{proof}
    The first equality is trivial. For the last two equalities, define \(x^\ast_i\) and \(y^\ast_i\) such that \(\imp(A,B)(x^{\ast}_i)\) is the \(i\)th largest value of \(\imp(A,B)(x)\) and \(A(y^\ast_i)\) is the \(i\)th smallest value of \(A(x)\) for all \(x\in X\) and \(i \in \{1,\dots,\abs{X}=n\}\). Now define \(k\) as follows:
    \[k := \abs{X \backslash A_{>0}}= \abs{\left\{x | A(x)=0\right\}},\]
    we then have, w.l.o.g., that
    \[\left\{x^\ast_1, \dots, x^\ast_k\right\}\subseteq X \backslash A_{>0},\]
    because of the fact that \(\imp(0, y) = 1\) for every implicator \(\imp\) and \(y \in [0,1]\).
    Looking at Equation \eqref{eq: yager binary quantifier} and realizing that \(A(y^\ast_{k+1})\) is the first non-zero \(A(y^\ast_i)\), we get that
    \[\widetilde{Y}^2_{\Lambda_\exists}(A,B)=\imp (A,B)(x^\ast_{k+1})=\max_{x\in A_{>0}}\left(\imp (A,B)(x) \right).\]
    The equality for \(\widetilde{W}^{\imp}_{\Lambda_\exists}(A,B)\) follows analogously.
\end{proof}
So, from this proposition we see that the \(\widetilde{Y}^2_{\Lambda_\exists}\) and \(\widetilde{W}^{\imp}_{\Lambda_\exists}\) models act more reasonably than \(\widetilde{Y}^{\imp}_{\Lambda_\exists}\), since for the former, the elements fully outside \(A\) do not influence the result. However, the most intuitive and simple evaluation of the binary existential quantifier is achieved by standard DFS models based on \(Q^2_{\Lambda_\exists}\).

\begin{prop}{\cite{glockner2008fuzzy}}
    For any standard DFS \(\mathcal{F}\), and \(A,B \in \fpwset(X)\), we have
    \[\mathcal{F}\left(Q^2_{\Lambda_\exists}\right)(A,B)= \max_{x\in X}\left((A\cap B)(x)\right).\]
\end{prop}

\subsubsection{Argument monotonicity}
In this subsection, we consider the properties of set monotonicity and relation monotonicity \cite{d2015comprehensive}, which are relevant to many applications of fuzzy rough sets. Note that all discussed binary quantification models are non-decreasing in the second argument, hence all FQFRS based on them satisfy set monotonicity for the lower approximation, i.e., the lower approximation is non-decreasing in the concept:
\[(\forall A,B \in \fpwset(X))\left(A\subseteq B \implies(\underline{apr}_{R,\widetilde{Q}_l}A) \subseteq (\underline{apr}_{R,\widetilde{Q}_l}B)\right).\]
In order to satisfy relation monotonicity for the lower approximation, i.e.,
\[(\forall R_1,R_2 \in \fpwset(X\times X))\left(R_1\subseteq R_2 \implies(\underline{apr}_{R_2,\widetilde{Q}_l}A) \subseteq (\underline{apr}_{R_1,\widetilde{Q}_l}A)\right),\] 
\(\widetilde{Q}_l\) should be increasing in its first argument. However, this only holds for fuzzy quantifiers that have \(Q^\rightarrow _\Lambda\) as their underlying semi-fuzzy quantifier, like \(\tilde{Y}^I_\Lambda\). Indeed, all models that correctly (i.e., intuitively) evaluate the sentence ``\(\Lambda\) \(A\)'s are \(B\)'s'' have, as we have seen in Section \ref{sec: problem}, \(Q^2_\Lambda\) as their underlying semi-fuzzy quantifier. But \(Q^2_\Lambda\) is not non-increasing in the first argument, since adding an element \(x\) to \(A\) can either increase or decrease the truth value, depending on whether the element is contained in \(B\) or not. This causes FQFRS models that are based on quantifiers with \(Q^2_\Lambda\) as their underlying semi-fuzzy quantifier to not satisfy relation monotonicity for their lower approximation.
 Applications where relation monotonicity is required for the lower approximation, as for example fuzzy-rough attribute reduction based on the QuickReduct algorithm \cite{cornelis2008noise}, should thus use OWAFRS.
\subsubsection{Inequalities between the different models}

The following proposition shows that the sentence ``For most \(X\)'s, if they are in \(A\), they are in \(B\)'' is always at least as true as the sentence ``Most \(A\)'s are \(B\)'s'', when restricting ourselves to crisp arguments.

\begin{prop}
    \label{prop: lowerapproxinequality}
    We have the following inequality:
    \[Q^{\rightarrow}_\Lambda(A,B)=Q^2_\Lambda(X, A\rightarrow B)\geq Q^2_\Lambda(A,B),\]
    for every \(A,B\in\pwset(X)\).
\end{prop}

\begin{proof}
    \begin{alignat*}{2}
         &      & \Lambda\left(\frac{\abs{\lnot A}+\abs{A\cap B}}{\abs{X}}\right) & \geq  \Lambda\left(\frac{\abs{A\cap B}}{\abs{A}}\right) \\
         & \iff & \frac{\abs{\lnot A}+\abs{A\cap B}}{\abs{\lnot A}+\abs{A}}       & \geq  \frac{\abs{A\cap B}}{\abs{A}}                     \\
         & \iff & (\abs{\lnot A}+\abs{A\cap B})*\abs{A}                           & \geq \abs{A\cap B}*(\abs{\lnot A}+\abs{A})              \\
         & \iff & \abs{\lnot A}*\abs{A}                                           & \geq \abs{A\cap B}*\abs{\lnot A}                        \\
         & \iff & \abs{A}                                                         & \geq \abs{A\cap B}
    \end{alignat*}
\end{proof}
We now ask ourselves if the previous inequality generalizes to fuzzy arguments. When using a plausible binary quantification model we would expect that it does. Note that all three binary quantification models \(\mathcal{F}_{owa}(Q^2_\Lambda)\), \(\widetilde{W}^{\imp}_\Lambda\) and \(\widetilde{Y}^{2}_\Lambda\) satisfy
\[\mathcal{F}_{owa}(Q^2_\Lambda)(X, A)= \widetilde{W}^{\imp}_\Lambda (X, A) = \widetilde{Y}^{2}_\Lambda(X,A)= \widetilde{Y}_\Lambda(A),\]
so extending Proposition \ref{prop: lowerapproxinequality} to fuzzy quantifiers yields that all of the proposed binary quantification models should be smaller than \(\widetilde{Y}^{\imp}_\Lambda\).
\begin{prop}
    \label{prop: ineqDFS}
    We have the following inequality for every DFS \(\mathcal{F}\):
    \[\mathcal{F}(Q^{\rightarrow}_\Lambda)\geq \mathcal{F}(Q^2_\Lambda).\]
\end{prop}
\begin{proof}
    Every DFS satisfies quantifier monotonicity \cite{glockner2008fuzzy}.
\end{proof}
\begin{cor}
    We have the following inequality:
    \[\widetilde{Y}^{\imp_{KD}}_\Lambda\geq \mathcal{F}_{owa}(Q^2_\Lambda).\]
\end{cor}
\begin{proof}
    Follows from Corollary \ref{cor: FOWA YAGER}.
\end{proof}
As the following proposition shows, the YWI model also preserves this important inequality.
\begin{prop}
    We have the following inequality:
    \[\widetilde{Y}^{\imp}_\Lambda(A,B)\geq \widetilde{Y}^2_{\Lambda}(A,B),\]
    for every \(A,B\in\fpwset(X)\).
\end{prop}

\begin{proof}
    We will prove that \(\mu_{\Lambda}\geq \mu'_\Lambda\), from which the proposition follows. Since \(\Lambda\) is monotone, it is sufficient to prove:
    \begin{align*}
             & \frac{i}{n} \geq \frac{\sum_{j=1}^i A(x^\ast_j)}{\sum_{j=1}^n A(x^{\ast}_{j})}                                                          \\
        \iff & i\cdot\sum_{j=1}^n A(x^{\ast}_{j}) \geq n \cdot \sum_{j=1}^i A(x^\ast_j)                                                                \\
        \iff & i\cdot\sum_{j=1}^n A(x^{\ast}_{j}) \geq (n-i) \cdot \sum_{j=1}^i A(x^\ast_j)+i\cdot \sum_{j=1}^i A(x^\ast_j)                            \\
        \iff & i\cdot\left(\sum_{j=1}^n A(x^{\ast}_{j})-\sum_{j=1}^i A(x^\ast_j)  \right)\geq (n-i)\cdot  \sum_{j=1}^i A(x^\ast_j)                     \\
        \iff & i\cdot\left(\underbrace{A(x^{\ast}_n) +\cdots+A(x^{\ast}_{i+1})}_{n-i\text{ elements}}\right)\geq (n-i) \cdot \sum_{j=1}^i A(x^\ast_j),
    \end{align*}
    and because \(A(x^{\ast}_j)\) is the \(j\)th smallest \(A(x)\) for every \(j\) this is indeed true:
    \begin{align*}
        i\cdot\left(\underbrace{A(x^{\ast}_n) +\cdots+A(x^{\ast}_{i+1})}_{n-i\text{ elements}}\right) & \geq  (n-i)\cdot i\cdot A(x^{\ast}_{i+1})
        \\ &\geq (n-i) \cdot \sum_{j=1}^i A(x^\ast_j).
    \end{align*}
\end{proof}

Unfortunately, \(\widetilde{W}^{\imp}_\Lambda\) does not satisfy the generalization of Proposition \ref{prop: lowerapproxinequality}, as the following example shows.

\begin{example}
    We will show that the following inequality does not hold:
    \[\widetilde{Y}^{\imp}_\Lambda(A,B)\geq \widetilde{W}^{\imp}_\Lambda(A,B),\]
    by evaluating
    \[\max_{A,B\in\fpwset(X)}\left(\widetilde{W}^{\imp}_{\id}(A,B)-\widetilde{Y}^{\imp}_{\id}(A,B)\right),\]
    for \(X=\{x_1,x_2\}\) (i.e., \(n=2\)), \(\imp=\imp_{KD}\) and \(\id(x)=x\) for every \(x\in[0,1]\). To gain insight, we first rewrite the difference between the two quantifiers as follows:
    \begin{align*}
        \Delta(A,B) & := \widetilde{W}^{\imp}_{\id}(A,B)-\widetilde{Y}^{\imp}_{\id}(A,B)                                              \\
                    & =\sum_{i=1}^{2}\left(\frac{\imp(A,B)(x_i)}{2}\right)\cdot \left(\frac{2 \cdot A(x_i)}{A(x_1)+ A(x_2)}-1\right).
    \end{align*}
    Now, assume \(B = \{x_1\}\) and \(A= \{(x_1, 1), (x_2, a)\}\) for some \(a\in [0,1]\). Using \(\imp(A,B)= \{(x_1, 1),(x_2,1-a)\}\), we get:
    \begin{align*}
        \Delta(A,B)                          & = \frac{a(1-a)}{2(a+1)}       \\
        \frac{\diff }{\diff a} (\Delta(A,B)) & = -\frac{a^2+2a-1}{2(a+1)^2}.
    \end{align*}
    Setting \(\frac{\diff }{\diff a} (\Delta(A,B))\) to zero we get that the maximal positive value of \(\Delta(A,B)\) is reached for \(a= \frac{-2+\sqrt{8}}{2}\):
    \[\max_{A,B}\left(\widetilde{W}^{\imp}_{\id}(A,B)-\widetilde{Y}^{\imp}_{\id}(A,B)\right) \approx 0.0858,\]
    for \(n=2\).
\end{example}
Finally, we have the following inequality that shows that the WOWA model always results in larger lower approximations than the YWI model.
\begin{prop}
    We have the following inequality:
    \[\widetilde{W}^{\imp}_\Lambda(A,B)\geq \widetilde{Y}^2_{\Lambda}(A,B),\]
for every \(A,B\in\fpwset(X)\).
\end{prop}
\begin{proof}
    Follows from \(\mu^A_{\Lambda}\geq \mu'_\Lambda\).
\end{proof}
\section{Experimental evaluation}
\label{sec: Experiment}
In this section, we evaluate the different FQFRS models based on different binary quantification strategies when applied to classification. Section \ref{subsec: classification} describes a simple classification algorithm that we will use to test the different FQFRS models. In Section \ref{subsec: setup}, we lay out the setup of the experiment, while Section \ref{subsec: results} discusses the results.
\subsection{Classification using fuzzy rough sets}
\label{subsec: classification}
The goal of classification is to predict the class of an instance, given a set of examples. More specifically, the set of examples is given in the form of a decision system \((X,\mathcal{A}\cup\{d\})\). For classification, we assume that \(d\) is a categorical attribute; the attributes of \(\mathcal{A}\) can either be categorical or numerical. The problem of classification is then to predict for a new instance \(x\notin X\), for which the evaluations of the conditional attributes are given, the value of \(d(x)\). A simple algorithm for classification \cite{vluymans2019dealing}, using fuzzy rough sets, is to classify a test instance to the decision class for which it has the greatest membership to the lower approximation of that class. In case of ties the algorithm chooses the first class. This algorithm is well-suited to test the different FQFRS models, since we have only focused on improving the lower approximation. To calculate these lower approximations, we need a fuzzy relation \(R\in\mathcal{F}(X\times X)\) describing the similarity between instances based on the conditional attributes. In the experiment below we will make use of the following relation:
\begin{equation*}
    R(x,y)=\frac{1}{|\mathcal{A}|}\sum_{a\in\mathcal{A}}R_a(x,y),
\end{equation*}
where
\[R_a(x,y)=\max\left(1-\frac{\abs{a(y)-a(x)}}{\sigma_a},0\right),\]
and \(\sigma_a\) denotes the standard deviation of \(a\).

\subsection{Setup}
\label{subsec: setup}
The different binary quantification models we will evaluate are:
\begin{itemize}
    \item OWA: OWA-based fuzzy rough sets, i.e., \(\widetilde{Y}^{\imp}_{\Lambda}\),
    \item WOWA: lower approximation calculated using \(\widetilde{W}^{\imp}_{\Lambda}\),
    \item YWI: lower approximation calculated using \(\widetilde{Y}^{2}_{\Lambda}\),
    \item FOWA: lower approximation calculated using \(\mathcal{F}_{OWA}\left(Q^2_{\Lambda}\right)\),
    \item VQFRS: vaguely quantified fuzzy rough sets, i.e., \(\widetilde{Z}^{2}_{\Lambda}\),
    \item \(\mathcal{M}_{CX}\): lower approximation calculated using \(\mathcal{M}_{CX}\left(Q^2_{\Lambda}\right)\),
    \item FRS: classical fuzzy rough sets, i.e., lower approximation calculated using the RIM quantifier \(\Lambda_\forall\) (cf. Proposition \ref{universalALL}).
\end{itemize}
The Kleene-Dienes implicator is used in all models. Notice that the exact choice of implicator is not important. Indeed, since the second argument is always crisp, only the choice of the underlying negator is important, which in our case is the standard negator.
The lower approximations will be evaluated using the RIM quantifiers \(\Lambda(x) = \Lambda_{(a,1)}(x)\) (Eq.\ \eqref{ZADEH_S_function}) for
\[a\in [0,0.1,0.2,0.3,0.4,0.5,0.6,0.7,0.8,0.9,0.925,0.95,0.975,0.99,0.999],\]
where we have chosen finer steps at the end to observe the convergence of \(\Lambda_{(a,1)}\) to the universal quantifier \(\Lambda_\forall\) as \(a\) approaches \(1\).
We evaluate the performance on 24 datasets (Table \ref{dataset_description}) from the UCI-repository \cite{Dua:2019} by means of stratified 5-fold cross-validation. All of the datasets only have numerical features. Since we are comparing noise-tolerant fuzzy rough set models, we add \(20\%\) class label noise to the datasets, i.e., for \(20\%\) randomly chosen elements we replace the class label with another randomly chosen class label. The balanced accuracy is used as the performance measure.
\begin{table}[H]
    \begin{center}
        \footnotesize
        \begin{tabular}{l| c c c || l | c c c}
            Name     & \# Cl. & \# Ft. & \# Inst.  & Name         & \# Cl. & \# Ft. & \# Inst.   \\
            \hline
            accent   & 6      & 12     & 329    & pop-failures & 2      & 18     & 540     \\
            \hline
            append.  & 2      & 7      & 106        & segment      & 7      & 19     & 2310      \\
            \hline
            banknote & 2      & 4      & 1372      & somerville   & 2      & 6      & 143       \\
            \hline
            biodeg   & 2      & 41     & 1055      & sonar        & 2      & 60     & 208       \\
            \hline
            breast.  & 6      & 9      & 106       & spectf       & 2      & 44     & 267        \\
            \hline
            coimbra  & 2      & 9      & 116       & sports.      & 2      & 59     & 1000       \\
            \hline
            debrecen & 2      & 19     & 1151      & transfusion  & 2      & 4      & 748        \\
            \hline
            faults   & 7      & 27     & 1941      & wdbc         & 2      & 30     & 569        \\
            \hline
            haber.   & 2      & 3      & 306       & wifi         & 4      & 7      & 2000      \\
            \hline
            ilpd     & 2      & 10     & 579       & wisconsin    & 2      & 9      & 683       \\
            \hline
            iono.    & 2      & 34     & 351       & wpbc         & 2      & 32     & 138       \\
            \hline
            leaf     & 30     & 14     & 340      & yeast        & 10     & 8      & 1484     \\
            \hline
        \end{tabular}
    \end{center}
    \caption{Description of the 24 used UCI datasets (\# Cl. = number of classes, \# Ft.\ = number of features, \# Inst.\ = number of instances).}
    \label{dataset_description}
\end{table}
\subsection{Results and discussion}
\label{subsec: results}
\begin{figure}[H]
    \centering
    \includegraphics[width=15cm]{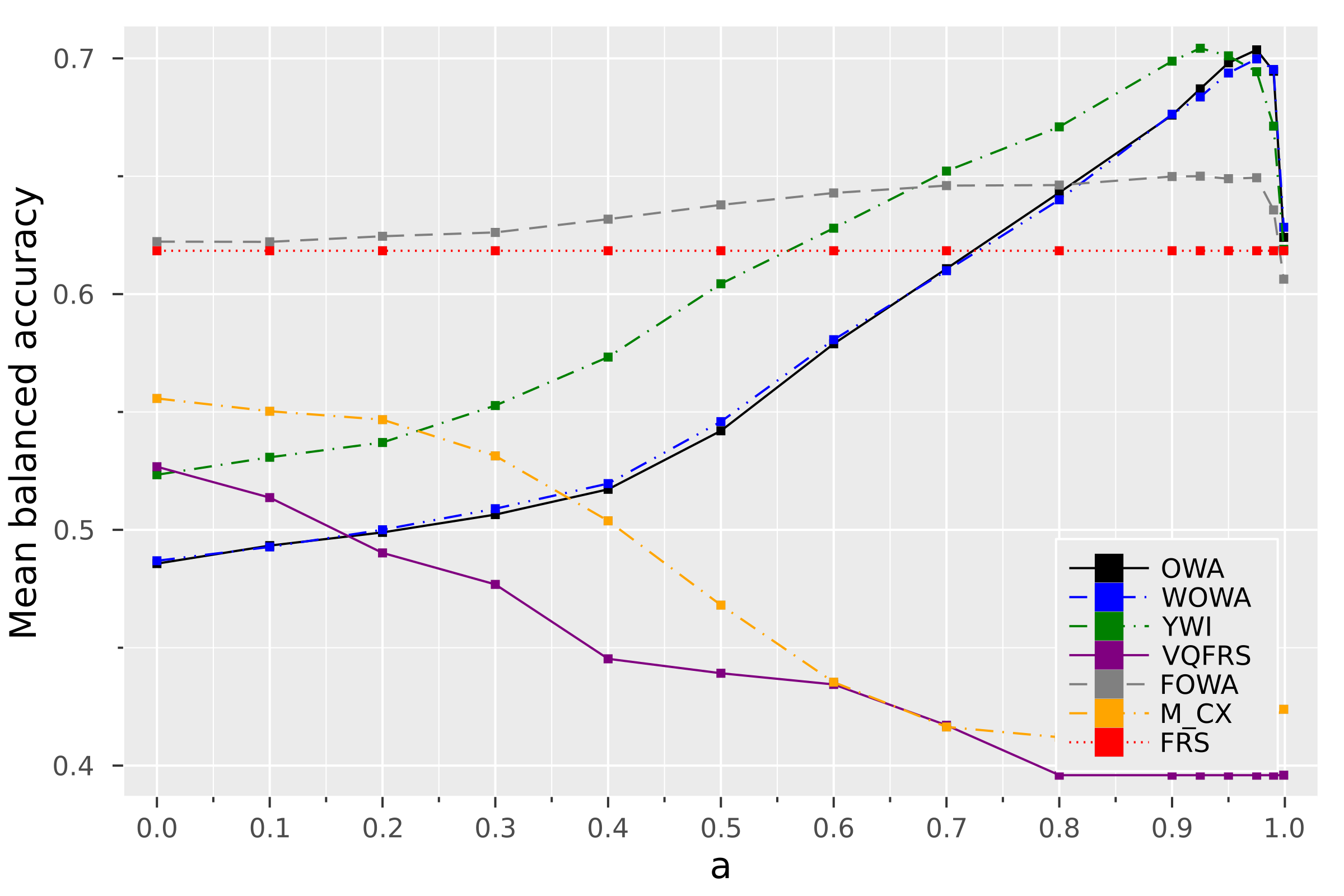}
    \caption{Plot of the mean balanced accuracy.}
    \label{balanced_accuracy_plot}
\end{figure}

\begin{figure}[H]
    \centering
    \includegraphics[width=15cm]{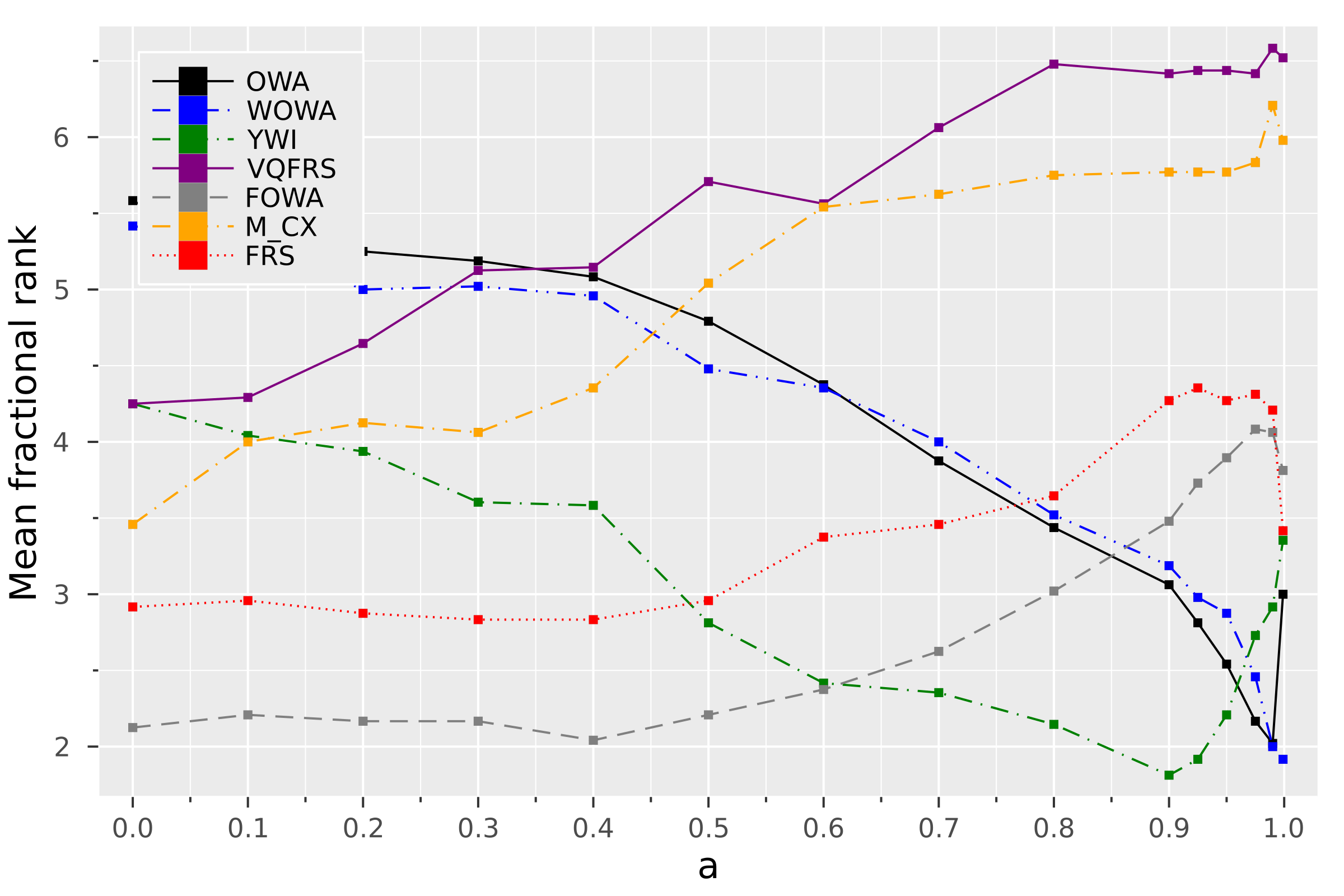}
    \caption{Plot of the mean fractional rank.}
    \label{rank_plot}
\end{figure}

\begin{figure}[H]
    \centering
    \includegraphics[width=15cm]{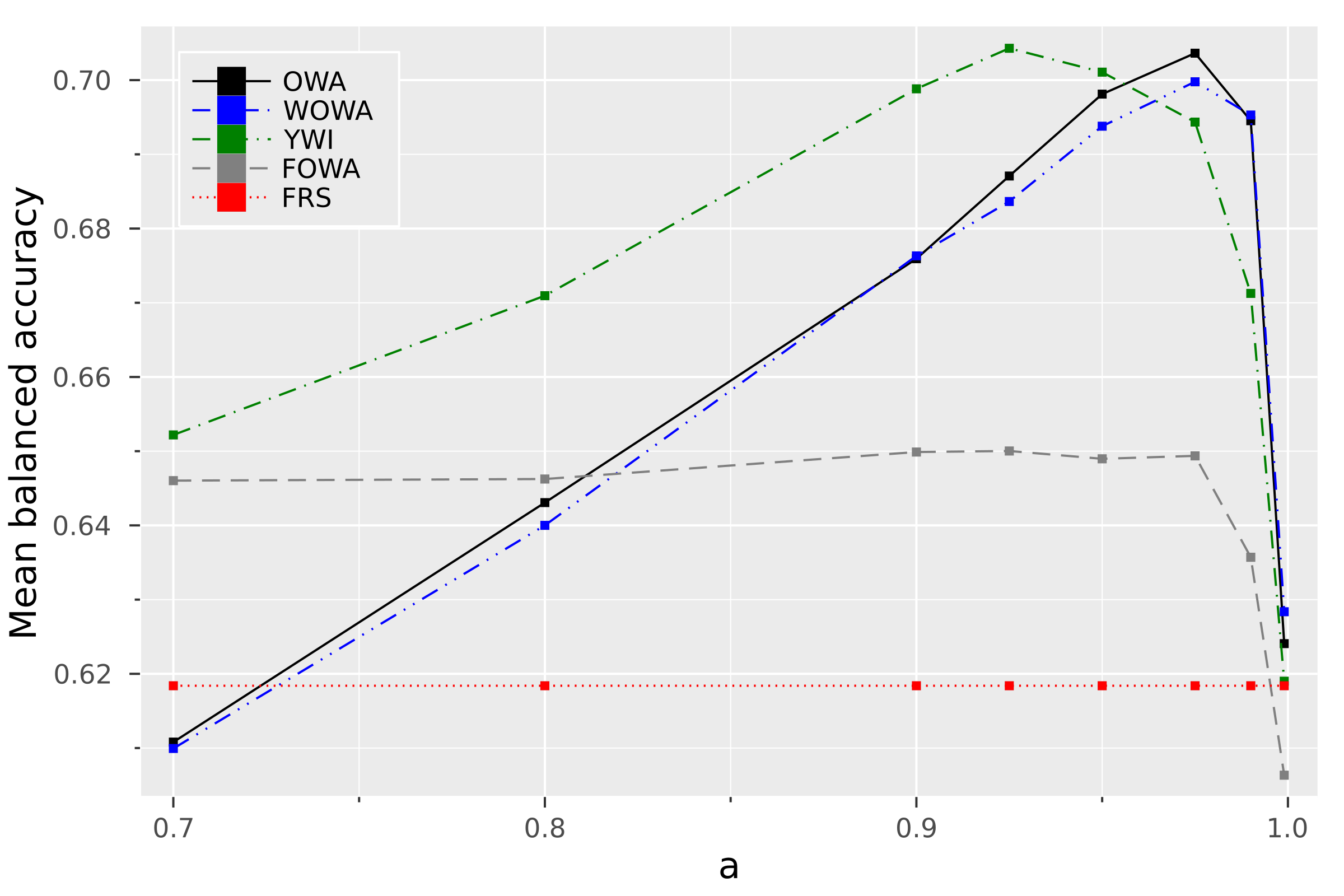}
    \caption{Zoomed in plot of the mean balanced accuracy.}
    \label{balanced_accuracy_plot_ZOOM}
\end{figure}

\begin{figure}[H]
    \centering
    \includegraphics[width=16cm]{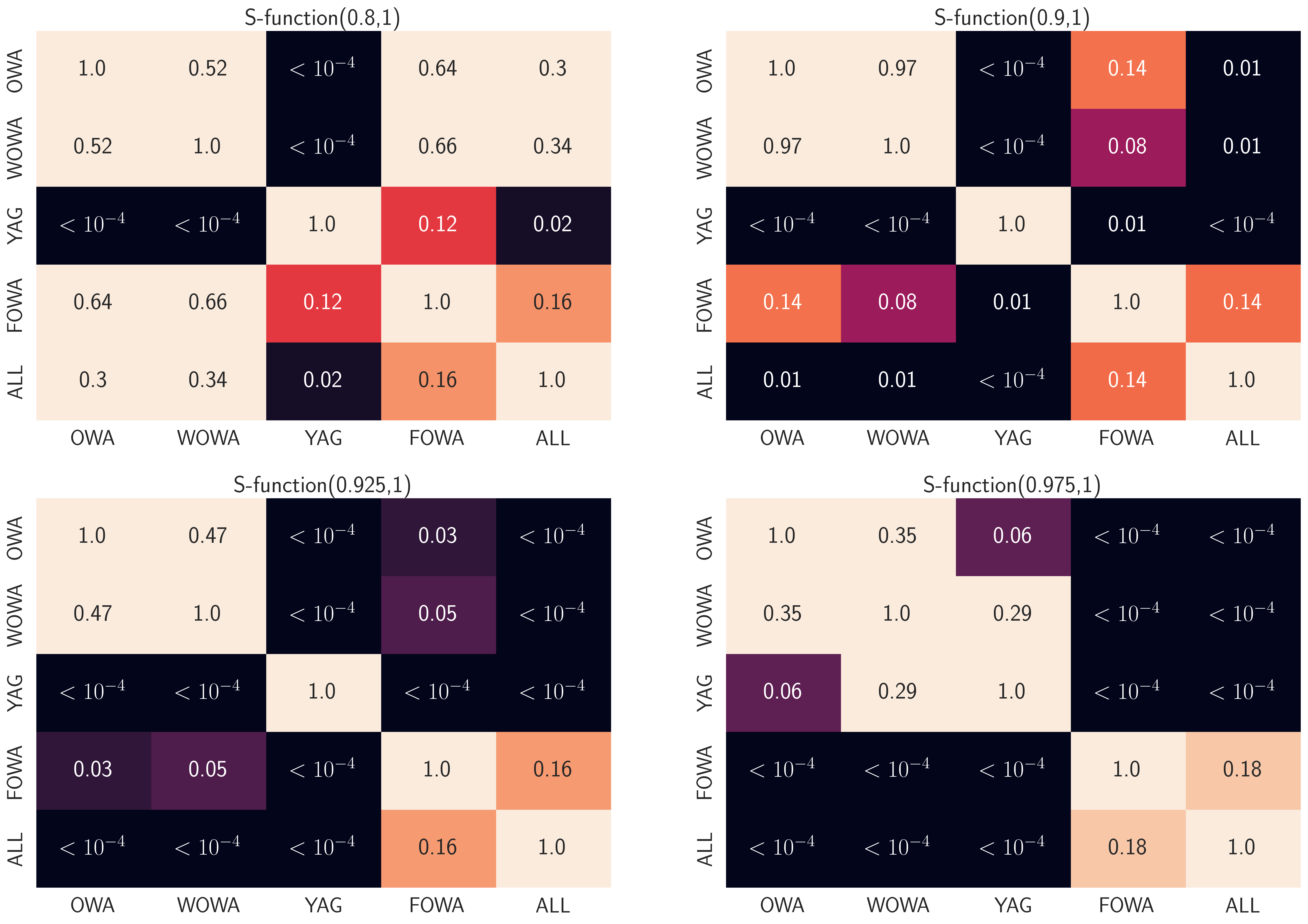}
    \caption{Heatmap of the \(p\)-values from the pairwise two-sided Wilcoxon signed rank test.}
    \label{heatmapBench}
\end{figure}

The mean balanced accuracy and mean fractional rank (i.e., items with equal scores receive the same ranking number, which is the mean of what they would have under ordinal rankings) are plotted against the RIM quantifier parameter \(a\) in Figures \ref{balanced_accuracy_plot} and \ref{rank_plot}, respectively. A zoomed-in plot of the mean balanced accuracy is given in Figure \ref{balanced_accuracy_plot_ZOOM}. The first thing we notice is that ZAD and \(\mathcal{M}_{CX}\) perform very poorly (mean balanced accuracy of less than 0.5).
An explanation for the poor performance of ZAD is that the ratio
\[\frac{\abs{Ry\cap C}}{\abs{Ry}}=\frac{\sum_{x\in C}R(x,y)}{\sum_{x\in X}R(x,y)},\]
where \(C\) is one of the decision classes, is generally very small and thus the lower approximations of ZAD are usually zero, which results in always choosing the first class and yielding a balanced accuracy of less than \(0.5\). The problem with the \(\mathcal{M}_{CX}\) model is exactly the problem that Glöckner addresses, i.e., the model is too cautious and evaluates sentences as \(0.5\) most of the time, thus resulting in the fact that many classes have the same value for their lower approximation.
Furthermore, we observe from Figures \ref{balanced_accuracy_plot} and \ref{rank_plot} that WOWA and OWA coincide almost everywhere, and thus that WOWA is not worth the extra complexity. A second observation is that the YWI quantifier almost always outperforms OWA and WOWA, except for \(a\) values close to \(1\), where OWA and WOWA perform the best. However, this worse performance for larger values of \(a\) can actually be seen as a good property of YWI, since it shows that YWI behaves more smoothly with respect to the RIM quantifiers. Indeed, when the \(a\) parameter approaches \(1\), the RIM quantifier \(\Lambda_{(a,1)}\) approaches the universal RIM quantifier \(\Lambda_\forall\), and thus, using Proposition \ref{universalALL} and \ref{UniveralALLGlock}, all of the lower approximations (except ZAD) should, from an intuitive perspective, approach the classical lower approximation. In Figure \ref{balanced_accuracy_plot}, we see that YWI and FOWA do this smoothly, as opposed to OWA and WOWA. Also note the overall robustness of FOWA with respect to the parameter \(a\), and that FOWA outperforms all other quantifiers for \(a\) values smaller than \(0.6\), but that the highest achieved performance is less than YWI, OWA and WOWA. So FOWA does have some benefits, however since it does not achieve a high maximum accuracy, it is not worth the extra complexity. Furthermore, note that YWI and OWA's highest achieved mean balanced accuracy are equal, but YWI sustains it for a wider range of $a$ values. In other words, YWI is more robust with respect to the RIM quantifier.
To discern if some of these methods outperform others consistently and significantly, we perform two-sided Wilcoxon signed ranks tests. The results of these tests are displayed in Figure \ref{heatmapBench}. Notice that YWI outperforms OWA and WOWA with very high significance (\(p < 10^{-4}\)) for \(a\) values smaller than \(0.95\). For \(a\)-values larger than \(0.95\) there is weak evidence (\(p < 0.1\)) that OWA outperforms YWI, agreeing with our conclusions above. Summarizing, these results show that YWI is the preferable method since it outperforms OWA for most \(a\) values and is smoother, which is desirable when dealing with hyperparameter optimization, while still achieving the highest possible accuracy of all the methods.

\section{Conclusion}
\label{sec: Conclusion}
We have introduced \emph{fuzzy quantifier-based fuzzy rough sets} (FQFRS), a general definition of fuzzy rough sets based on fuzzy quantifiers. FQFRS allows to position existing models and compare them on the basis of their associated fuzzy quantifiers. In addition, this general model can lead to improved models in terms of performance and interpretability by using semantically more sound binary quantification models. Furthermore, we have introduced different binary quantification models that can be used with FQFRS and studied some of their theoretical properties. From this theoretical analysis, we can conclude that the YWI binary quantification model (YWI-FQFRS), together with the DFS-based models FOWA and \(\mathcal{M}_{CX}\), behave the most intuitively. In addition, we have demonstrated that YWI-FQFRS acts more smoothly with respect to the RIM quantifier than OWAFRS, aiding hyperparameter tuning, and overall outperforms OWAFRS when choosing the RIM quantifier a priori. All these benefits make YWI-FQFRS a worthy competitor to OWAFRS. Finally, we have shown that on every level YWI-FQFRS is an improvement on VQFRS.

One direction for future research is to find out how the properties of the used quantifiers translate to properties of the corresponding fuzzy rough sets, and vice versa. A theoretical study of the smoothness with respect to the RIM quantifier would also be interesting. Furthermore, testing FQFRS on an application where both arguments are fuzzy, as well as experimenting with different similarity relations, is necessary to investigate the performance of FQFRS. Finally, proposing new binary quantification models for the lower approximation in FQFRS might yield even better fuzzy rough set models.

\section*{Acknowledgment}
The research reported in this paper was conducted with the financial support of the Odysseus programme of the Research
Foundation – Flanders (FWO). The grant number is G0H9118N.

\bibliography{/Users/adnantheerens/Desktop/phd/Articles/bibfiles/bibfile.bib}
\end{document}